\documentclass{article}

\PassOptionsToPackage{numbers, compress}{natbib}

\usepackage[utf8]{inputenc}
\usepackage[T1]{fontenc}
\usepackage[draft]{hyperref}
\usepackage{url}
\usepackage{booktabs}
\usepackage{amsfonts}
\usepackage{nicefrac}
\usepackage{microtype}

\usepackage{graphicx}
\usepackage{amsmath}
\usepackage{amssymb}
\usepackage{bm}
\usepackage{algorithm}
\usepackage{algorithmic}
\usepackage{subcaption}

\usepackage{wrapfig}

\usepackage{color}

\usepackage[final]{neurips_2019}

\newcommand{\Cov}{\mathrm{Cov}}

\title{Function-Space Distributions over Kernels}

\author{
  \parbox{\linewidth}{
    \centering
    Gregory W. Benton\thanks{Equal contribution. $\ddagger$Work done during internship with AGW.}\, $^1$\quad
    Wesley J. Maddox$^{*2}$ \quad
    Jayson P. Salkey$^{*1}$\quad \\
    J\'ulio Albinati$^\ddagger$$^3$\quad
    Andrew Gordon Wilson$^{1,2}$
  }\\
  ~\\
  \parbox{\linewidth}{
    \centering
    $^1$Courant Institute of Mathematical Sciences, New York University\\
    $^2$Center for Data Science, New York University\\
    $^3$Microsoft\\
  }
}

\begin{document}

\maketitle

\begin{abstract}

Gaussian processes are flexible function approximators, with inductive biases controlled by a covariance kernel.
Learning the kernel is the key to representation learning and strong predictive performance. In this paper, we develop \emph{functional kernel learning} (FKL) to directly infer
functional posteriors over kernels.
In particular, we place a transformed Gaussian process over a spectral density, to induce a
non-parametric distribution over kernel functions. The resulting approach enables learning of rich representations, with support for any stationary kernel, uncertainty over the values of the kernel, and an interpretable specification
of a prior directly over kernels, without requiring sophisticated initialization or manual intervention. We perform inference through elliptical slice sampling, which is especially well suited to marginalizing posteriors with the strongly correlated priors typical to function space modeling. We develop our approach for non-uniform, large-scale, multi-task, and multidimensional data, and show promising performance in a wide range of settings, including interpolation, extrapolation, and kernel recovery experiments.
\end{abstract}

\section{Introduction}
\label{intro}

Practitioners typically follow a two-step modeling procedure: (1) choosing the functional form of a model, such as a neural network; (2) focusing learning efforts on training the parameters of that model. While inference of these parameters consume our efforts, they are rarely interpretable, and are only of interest insomuch as they combine with the functional form of the model to make predictions. Gaussian processes (GPs) provide an alternative \emph{function space} approach to machine learning, directly placing a distribution over functions that could fit data \citep{gpml}. This approach enables great flexibility, and also provides a compelling framework for controlling the inductive biases of the model, such as whether we expect the solutions to be smooth, periodic, or have conditional independence properties.

These inductive biases, and thus the generalization properties of the GP, are determined by a kernel function. The performance of the GP, and what representations it can learn, therefore crucially depend on what we can learn about the kernel function itself. Accordingly, kernel functions are becoming increasingly expressive and parametrized \citep{jang2017scalable, tobar_learning_nodate, wilson2013gaussian}. There is, however, no a priori reason to assume that the true data generating process is driven by a particular parametric family of kernels.

\begin{figure}
	\centering
        \includegraphics[width=1\linewidth]{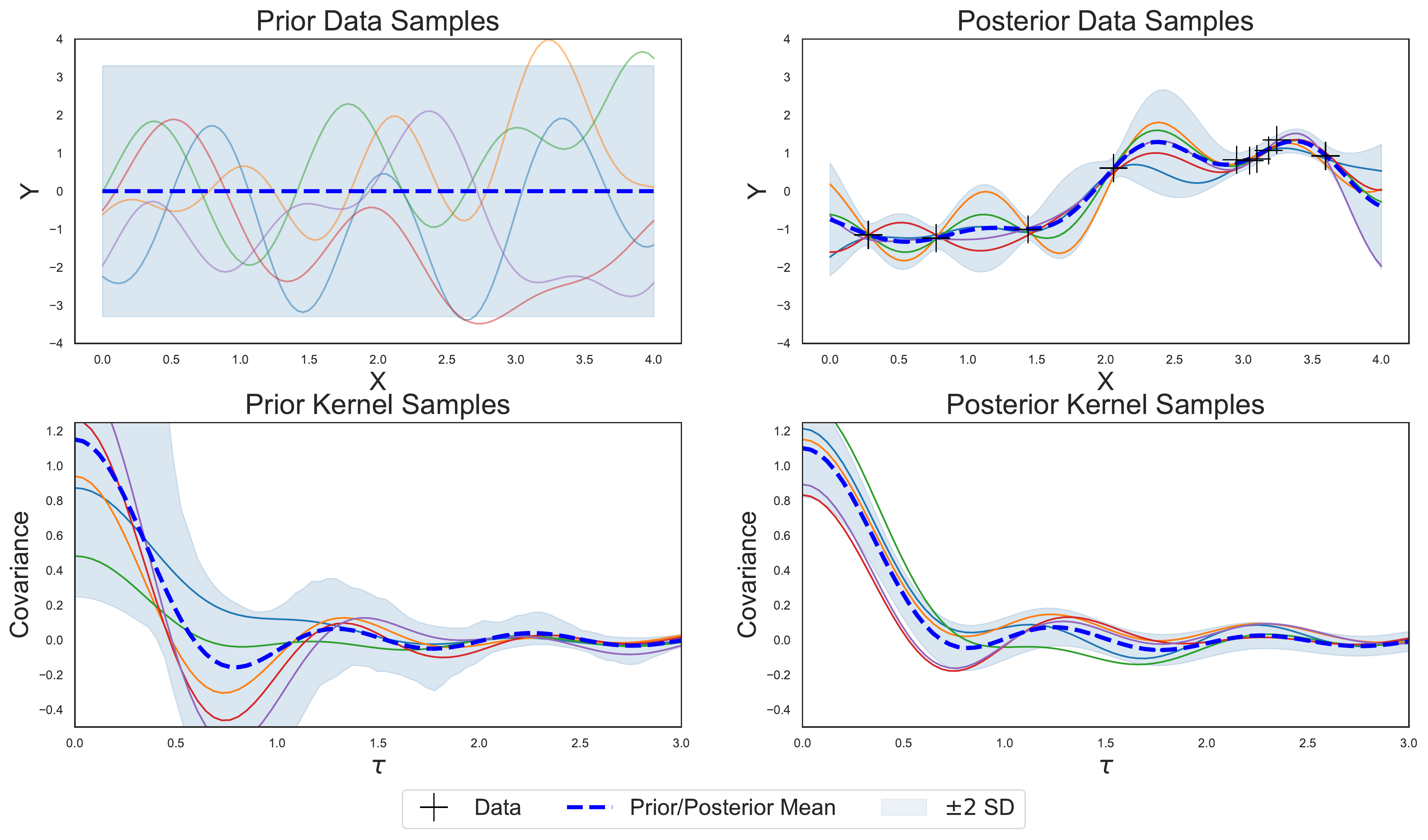}
       \caption{\textbf{Above:} A function-space view of regression on data. We show draws from a GP prior and posterior over functions in the left and right panels, respectively. \textbf{Below:} With FKL, we apply the function-space view to \emph{kernels}, showing prior kernel draws on the left, and posterior kernel draws on the right.
         In both cases, prior and posterior means are in thick black, two standard deviations about the mean in grey shade, and data points given by crosses. With FKL, one can specify the prior mean over kernels to be any parametric family, such an RBF kernel, to provide a useful \emph{inductive bias}, while still containing support for \emph{any} stationary kernel.}
       \label{fig: gp-kernels}
       \vspace{-0.5cm}
\end{figure}

We propose extending the function-space view to kernel learning itself -- to represent uncertainty over the kernel function, and to reflect the belief that the kernel does not have a simple parametric form. Just as one uses GPs to directly specify a prior and infer a posterior over functions that can fit data, we propose to directly reason about priors and posteriors over kernels.
In Figure \ref{fig: gp-kernels}, we illustrate the shift from standard function-space GP regression, to a function-space view of kernel learning.

Specifically, our contributions are as follows:
\begin{itemize}
  \item We model a spectral density as a transformed Gaussian process, providing a non-parametric function-space distribution over kernels. Our approach, \emph{functional kernel learning} (FKL), has several key properties: (1) it is highly flexible, with support for any stationary covariance function; (2) it naturally represents uncertainty over all values of the kernel; (3) it can easily be used to incorporate intuitions about what types of kernels are \emph{a priori} likely; (4) despite its flexibility, it does not require sophisticated initialization or manual intervention; (5) it provides a conceptually appealing approach to kernel learning, where we reason directly about prior and posterior kernels, rather than about parameters of these kernels.

  \item We further develop FKL to handle multidimensional and irregularly spaced data, and multi-task learning.

    \item We demonstrate the effectiveness of FKL in a wide range of settings, including interpolation, extrapolation, and kernel recovery experiments, demonstrating strong performance compared to state-of-the-art methods.

  \item Code is available at \url{https://github.com/wjmaddox/spectralgp} .
\end{itemize}

Our work is intended as a step towards developing Gaussian processes for \emph{representation learning}. By pursuing a function-space approach to kernel learning, we can discover rich representations of data, enabling strong predictive performance, and new interpretable insights into our modeling problems.

\section{Related Work}

We assume some familiarity with Gaussian processes \citep[e.g.,][]{gpml}.
A vast majority of kernels and kernel learning methods are parametric.
Popular kernels include the parametric RBF, Mat\'ern, and periodic kernels. The standard multiple kernel learning \citep{genton2001classes,gonen2011multiple,lanckriet2004learning,rakotomamonjy2007more}
approaches typically involve additive compositions of RBF kernels with different bandwidths. More recent methods model the spectral density (the Fourier transform) of stationary kernels to construct kernel learning procedures. \citet{lazaro-gredilla_sparse_nodate} models the spectrum as independent point masses. \citet{wilson2013gaussian} models the spectrum as a scale-location mixture of Gaussians, referred to as a \emph{spectral mixture kernel} (SM). \citet{yang2015carte} combine these approaches, using a random feature expansion for a spectral mixture kernel, for scalability. \citet{oliva2016bayesian} consider a Bayesian non-parametric extension of \citet{yang2015carte}, using a random feature expansion for a Dirichlet process mixture. Alternatively, \citet{jang2017scalable} model the parameters of a SM kernel with prior distributions, and infer the number of mixture components. While these approaches provide strong performance improvements over standard kernels, they often struggle with difficulty specifying a prior expectation over the value of the kernel, and multi-modal learning objectives, requiring sophisticated manual intervention and initialization procedures \citep{herlands2018change}.

A small collection of pioneering works \citep{tobar2018bayesian, tobar_learning_nodate, wilson2014covariance} have considered various approaches to modeling the spectral density of a kernel with a Gaussian process. Unlike FKL, these methods are constrained to one-dimensional time series, and still require significant intervention to achieve strong performance, such as choices of windows for convolutional kernels. Moreover, we demonstrate that even in this constrained setting, FKL provides improved performance over these state-of-the-art methods.

\section{Functional Kernel Learning}

In this section, we introduce the prior model for \emph{functional kernel learning} (FKL). FKL induces a distribution over kernels by
modeling a spectral density (Section \ref{spectral}) with a transformed Gaussian process (Section \ref{sec: model-spec}). Initially we consider one
dimensional inputs $x$ and outputs $y$, and then generalize the approach to multiple input dimensions (Section \ref{sec: multi-input}), and multiple output dimensions (multi-task) (Section \ref{sec: multi-task}). We consider inference within this model in Section \ref{sec: inference}.

\subsection{Spectral Transformations of Kernel Functions}
\label{spectral}

Bochner's Theorem \citep{bochner, gpml} specifies that $k(\cdot)$ is the covariance of a stationary process on $\mathbb{R}$ if and only if
\vspace{-0.3cm}
\begin{align}\label{eqn: fourier-trans}
k(\tau) = \int_{\mathbb{R}}e^{2\pi i \omega \tau}S(\omega) d\omega,
\end{align}
where $\tau = |x-x'|$ is the difference between any pair of inputs $x$ and $x'$,
for a positive, finite \emph{spectral density} $S(\omega)$.
 This relationship is reversible: if $S(\omega)$ is known, $k(\tau)$ can be computed via inverse Fourier transformation.

 \begin{wrapfigure}[19]{L}{0.33\textwidth}
 	\vspace{-0.2cm}
   \centering
   \includegraphics[width=\linewidth]{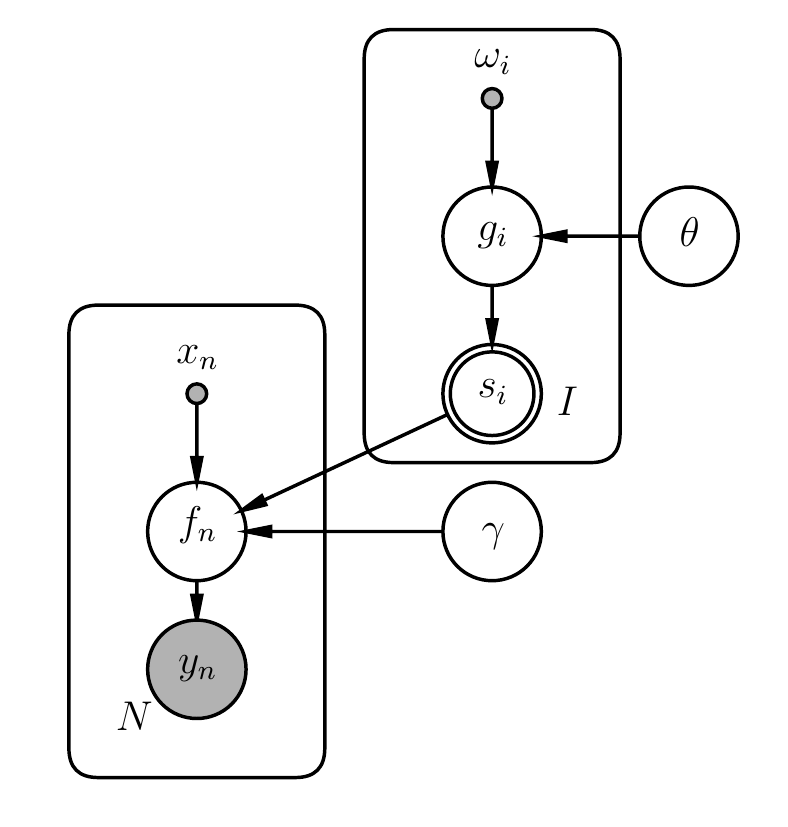}
   \caption{Graphical model for the FKL framework. Observed data is $y_n$, corresponding to the GP output $f_n$. The spectral density is $S_i$ for observed frequencies $\omega_i.$, and hyper-parameters are $\phi = \{\theta, \gamma\}.$}
   \label{fig:graphical_model}
 \end{wrapfigure}

For $k(\tau)$ to be real-valued, $S(\omega)$ must be symmetric. Furthermore, for finitely sampled $\tau$ we are only able to identify angular frequencies up to $2\pi/\Delta$ where $\Delta$ is the minimum absolute difference between any two inputs.
Equation \ref{eqn: fourier-trans} simplifies to
\begin{equation}\label{eqn: cos}
k(\tau) = \int_{[0, 2\pi / \Delta)} \cos(2\pi \tau \omega) S(\omega) d\omega,
\end{equation}
by expanding the complex exponential and using the oddness of sine (see Eqs. 4.7 and 4.8 in \citet{gpml}) and then truncating the integral to the point of identifiability.

For an arbitrary function, $S(\omega),$ Fourier inversion does not produce an analytic form for $k(\tau)$, however we can use simple numerical integration schemes like the trapezoid rule to approximate the integral in Equation \ref{eqn: cos} as

\begin{align}\label{eqn: trap}
k(\tau) \approx \frac{ \Delta_\omega}{2} \sum_{i=1}^{I} \cos(2\pi \tau \omega_i) S(\omega_i) +  \cos(2\pi \tau \omega_{i-1}) S(\omega_{i-1}),
\end{align}
where the spectrum is sampled at $I$ evenly spaced frequencies $\omega_i$ that are $\Delta_\omega$ units apart in the frequency domain.

The covariance $k(\tau)$ in Equation \eqref{eqn: trap} is periodic. In practice, frequencies can be chosen such that the period is beyond the bounds that would need to be evaluated in $\tau$.
As a simple heuristic we choose $P$ to be $8\tau_{max}$, where $\tau_{max}$ is the maximum distance between training inputs. We then choose frequencies so that $\omega_{n} = 2\pi n/P$ to ensure $k(\tau)$ is $P$-periodic. We have found choosing 100 frequencies ($n=0, \dots, 99$) in this way leads to good performance over a range of experiments in Section \ref{sec: experiments}.

\subsection{Specification of Latent Density Model}\label{sec: model-spec}
Uniqueness of the relationship in Equation \ref{eqn: fourier-trans} is guaranteed by the Wiener-Khintchine Theorem (see Eq. 4.6 of \citet{gpml}), thus learning the spectral density of a kernel is sufficient to learn the kernel.
We propose modeling the log-spectral density of kernels using GPs. The log-transformation ensures that the spectral representation is non-negative.
We let $\phi = \{\theta, \gamma\}$ be the set of \textit{all} hyper-parameters (including those in both the data, $\gamma,$ and latent spaces, $\theta$), to simplify the notation of Section \ref{sec: inference}.

\begin{figure}
    \includegraphics[width=\linewidth]{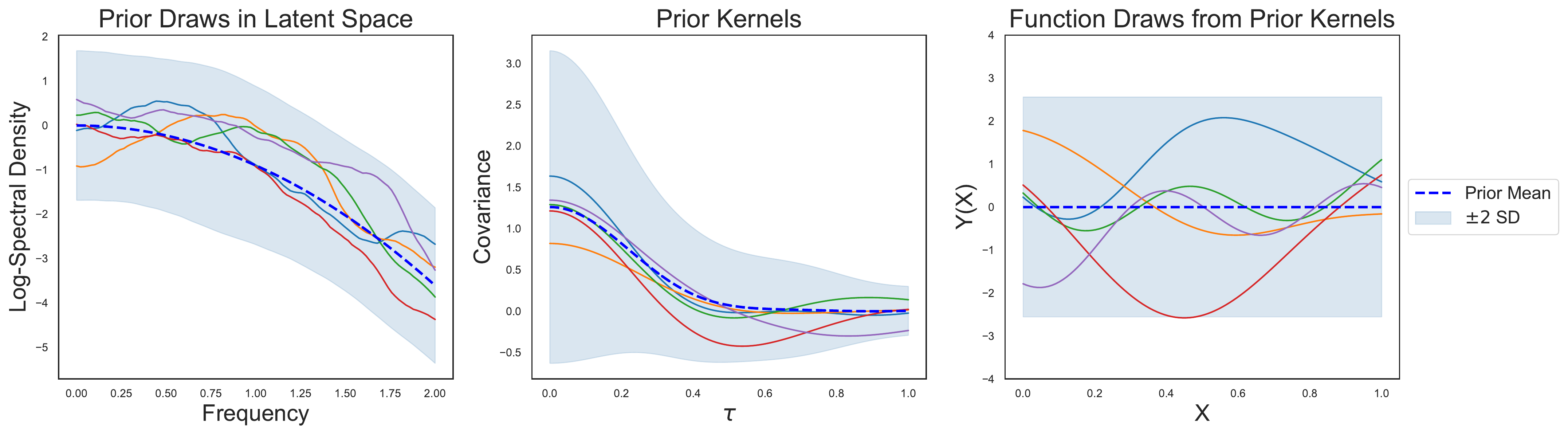}
    \caption{Forward sampling from the hierarchical FKL model of Equation \eqref{eq:model}. \textbf{Left}: Using randomly initialized hyper-parameters $\phi$, we draw functions $g(\omega)$ from the latent GP modeling the log spectral density. \textbf{Center}: We use the latent realizations of $g(\omega)$ with Bochner's Theorem and Eq.~\eqref{eqn: trap} to compose kernels. \textbf{Right}: We sample from a mean-zero Gaussian process with a kernel given by each of the kernel samples.
        Shaded regions show 2 standard deviations above and below the mean in dashed blue. Notice that the shapes of the prior kernel samples have significant variation but are clearly influenced by the prior mean, providing a controllable inductive bias.}
    \label{fig: prior-draws}
    \vspace{-0.6cm}
\end{figure}

Using Equation \ref{eqn: trap} to produce a kernel $k(\tau)$ through $S(\omega)$, the hierarchical model over the data is
\begin{equation}\label{eq:model}
\begin{aligned}
&\{\text{Hyperprior}\} \quad \hspace{2.76cm}p(\phi) = p(\theta, \gamma)\\
&\{\text{Latent GP}\} \qquad  \qquad \qquad \text{ }\text{ } \hspace{0.75cm}g(\omega) | \theta \sim \mathcal{GP}\left(\mu(\omega; \theta), k_{g}(\omega, \omega'; \theta)\right) \\
&\{\text{Spectral Density}\} \quad \hspace{1.90cm} S(\omega) = \exp\{g(\omega)\}\\
&\{\text{Data GP}\} \qquad \quad \text{ }\text{ }\text{ }\text{ } \hspace{0.75cm} f(x_n) | S(\omega), \gamma \sim \mathcal{GP}(\gamma_0, k(\tau; S(\omega))). \\
\end{aligned}
\end{equation}
We let $f(x)$ be a noise free function that forms part of an observation model. For regression, we can let $y(x) = f(x) + \epsilon(x)$, $\epsilon \sim \mathcal{N}(0,\alpha^2)$ (in future equations we implicitly condition on hyper-parameters of the noise model, e.g., $\alpha^2$, for succinctness, but learn these as part of $\phi$). The approach can easily be adapted to classification through a different observation model; e.g., $p(y(x)) = \sigma(y(x) f(x))$ for binary classification with labels $y \in \{-1,1\}$.
Full hyper-parameter prior specification is given in Appendix \ref{app:lm}. Note that unlike logistic Gaussian process density estimators \citep{adams_nonparametric_2009, tokdar2007posterior} we need not worry about the normalization factor of $S(\omega)$, since it is absorbed by the scale of the kernel over data, $k(0)$.
The hierarchical model in Equation \ref{eq:model} defines the functional kernel learning (FKL) prior, with corresponding graphical model in Figure \ref{fig:graphical_model}.
Figure \ref{fig: prior-draws} displays the hierarchical model, showing the connection between spectral and data spaces.

A compelling feature of FKL is the ability to conveniently specify a prior expectation for the kernel by specifying a mean function for $g(\omega)$, and to encode smoothness assumptions by the choice of covariance function.
For example, if we choose the mean of the latent process $g(\omega$) to be negative quadratic, then prior kernels are concentrated around RBF kernels, encoding the inductive bias that function values close in input space are likely to have high covariance.
In many cases the spectral density contains sharp peaks around dominant frequencies, so we choose a Mat\'ern $3/2$ kernel for the covariance of $g(\omega)$ to capture this behaviour.

\subsection{Multiple Input Dimensions}
\label{sec: multi-input}

We extend FKL to multiple input dimensions by either corresponding each one-dimensional kernel in a product of kernels with its own latent GP with distinct hyper-parameters (FKL separate) or having all one-dimensional kernels be draws from a single latent process with one set of hyper-parameters (FKL shared).
The hierarchical Bayesian model over the $d$ dimensions is described in the following manner:
\vspace{-0.2cm}
\begin{equation}\label{eq:multi_in_model}
\begin{aligned}
&\{\text{Hyperprior}\}\hspace{3.58cm}p(\phi) = p(\theta, \gamma) \\
&\{\text{Latent GP}\  \forall d \in \{1,... D\}\} \hspace{1cm} g_d(\omega_d) | \theta \sim \mathcal{GP}\left(\mu(\omega_d; \theta), k_{g_d}(\omega_d, \omega_d'; \theta)\right) \\
&\{\text{Product Kernel GP}\}\hspace{0.15cm} f(x) | \{g_d(\omega_d)\}_{d=1}^D, \gamma \sim \mathcal{GP}(\gamma_0, \prod_{d=1}^D k(\tau_d; S(\omega_d)))
\end{aligned}
\end{equation}

Tying the kernels over each dimension while considering their spectral densities to be draws from the same latent process (FKL shared) provides multiple benefits.  Under these assumptions, we have more information to learn the underlying latent GP $g(\omega)$.
We also have the helpful inductive bias that the covariance functions across each dimension have some shared high-order properties, and enables linear time scaling with dimensionality.

\subsection{Multiple Output Dimensions}\label{sec: multi-task}

FKL additionally provides a natural way to view multi-task GPs. We assume that each task (or output), indexed by
 $t \in \{ 1, \dots, T \}$,  is generated by a GP with a distinct kernel. The kernels are tied together by assuming each of those $T$ kernels are constructed from realizations of a single \emph{shared} latent GP.
Notationally, we let $g(\omega)$ denote the latent GP, and use subscripts $g_t(\omega)$ to indicate independent realizations of this latent GP. The hierarchical model can then be described in the following manner:
\begin{equation}\label{eq:mt_model}
\begin{aligned}
\{\text{Hyperprior}\} \hspace{3.58cm} p(\phi) &= p(\theta, \gamma) \\
\{\text{Latent GP}\}  \hspace{3.48cm} g(\omega) | \theta &\sim \mathcal{GP}\left(\mu(\omega; \theta), k_{g}(\omega, \omega'; \theta)\right) \\
\{\text{Task GP } \forall t \in \{1, ... T\}\} \hspace{0.56cm} f_{t}(x) | g_t(\omega), \gamma &\sim \mathcal{GP}(\gamma_{0, t}, k(\tau; S_t(\omega)))
\end{aligned}
\end{equation}
In this setup, rather than having to learn the kernel from a single realization of a process (a single task), we can learn the kernel
from multiple realizations, which provides a wealth of information for kernel learning \citep{wilson2015human}. While sharing individual hyper-parameters across multiple tasks is standard (see e.g. Section 9.2 of \citet{mackay1998introduction}), these approaches can only learn limited structure. The information provided by multiple tasks is distinctly amenable to FKL, which shares a flexible \emph{process over kernels} across tasks. FKL can use this information to discover unconventional structure in data, while retaining computational efficiency (see Appendix \ref{sec: scalability}).

\section{Inference and Prediction}\label{sec: inference}
When considering the hierarchical model defined in Equation \ref{eq:model}, one needs to learn both the hyper-parameters, $\phi$, and an instance of the latent Gaussian process, $g(\omega).$
We employ alternating updates in which the hyper-parameters $\phi$ and draws of the latent GP are updated separately.
A full description of the method is in Algorithm \ref{alg: alt-sampler} in Appendix \ref{app:lm}.
\paragraph{Updating Hyper-Parameters:}
Considering the model specification in Eq. \ref{eq:model}, we can define a loss as a function of $\phi = \{\theta, \gamma\}$ for an observation of the density, $\tilde{g}(\omega)$, and data observations $y(x)$. This loss corresponds to the entropy, marginal log-likelihood of the latent GP with fixed data GP, and the marginal log-likelihood of the data GP.
\begin{equation}\label{eqn: theta-loss}
\mathcal{L}(\phi) = - \left(\log p(\phi) + \log p(\tilde{g}(\omega)|\theta, \omega) + \log p(y(x) | \tilde{g}(\omega), \gamma, x)\right).
\end{equation}
This objective can be optimized using any procedure; we use the AMSGRAD variant of Adam as implemented in PyTorch \citep{reddi2019convergence}.
For GPs with $D$ input dimensions (and similarly for $D$ output dimensions), we extend Eq. \ref{eqn: theta-loss} as
\begin{equation}\label{eqn: mt-theta-loss}
\mathcal{L}(\phi) = - \left(\log p(\phi) + \sum_{d=1}^{D}\left[\log p(\tilde{g}_d(\omega_d)|\theta, \omega) \right]+ \log p(y(x) | \{\tilde{g}_d(\omega_d)\}_{d=1}^D, \gamma, x)\right).
\end{equation}

\paragraph{Updating Latent Gaussian Process:}
With fixed hyper-parameters $\phi$, the posterior of the latent GP is
\begin{equation}\label{eqn: g-loss}
p(g(\omega)|\phi, x, y(x), f(x)) \propto \mathcal{N}(\mu(\omega; \theta), k_g(\omega;\theta)) p(f(x) | g(\omega), \gamma).
\end{equation}
We sample from this posterior using elliptical slice sampling (ESS) \citep{murray2010elliptical,murray2010slice}, which is specifically designed to sample from posteriors with highly correlated Gaussian priors.
Note that we must reparametrize the prior by removing the mean before using ESS; we then consider it part of the likelihood afterwards.

Taken together, these two updates can be viewed as a single sample Monte Carlo expectation maximization (EM) algorithm \citep{wei_monte_1990} where only the final $g(\omega)$ sample is used in the Monte Carlo expectation.
Using the alternating updates (following Algorithm \ref{alg: alt-sampler}) and transforming the spectral densities into kernels, samples of predictions on the training and testing data can be taken.
We generate posterior estimates of kernels by fixing $\phi$ after updating and drawing samples from the posterior distribution, $p(g(\omega) | f, y, \phi)$, taken from ESS (using $y$ as short for $y(x)$, the training data indexed by inputs $x$).

\paragraph{Prediction:}
The predictive distribution for any test input $x^*$ is given by
\begin{align}
p(f^* | x^*, x, y, \phi) = \int p(f^* | x^*, x, y, \phi, k) p(k | x^*, x, y, \phi) dk
\end{align}
where we are only conditioning on data $x, y$, and hyper-parameters $\phi$ determined from optimization, by
\emph{marginalizing} the whole posterior distribution over kernels $k$ given by FKL. We use simple Monte Carlo to approximate
this integral as
\begin{align}
p(f^* | x^*, x, y, \phi) \approx \frac{1}{J} \sum_{j=1}^{J} p(f^* | x^*, x, y, \phi, k_j) \,, \quad k_j \sim p(k | x^*, x, y, \phi).
\end{align}
We sample from the posterior over $g(\omega)$ using elliptical slice sampling as above. We then transform these samples $S(\omega) = \exp\{g(\omega)\}$ to form posterior samples from the spectral density. We then sample $k_j \sim p(k | x^*, x, y, \phi)$ by evaluating the trapezoidal approximation in Eq.~\eqref{eqn: trap} (at a collection of frequencies $\omega$) for each sample of the spectral density. For regression with Gaussian noise $p(f^* | x^*, x, y, \phi, k)$ is Gaussian, and our expression for the predictive distribution becomes
\begin{equation}
\begin{aligned}\label{eq:pred}
p(f^* | x^*, x, y, \phi, \omega) &= \frac{1}{J} \sum_{j=1}^J \mathcal{N}(\bar{f}^*(x^*)_j , \Cov(f^*)_j) \\
\bar{f}^*(x^*)_j  &= k_{f_j}(x^*, x; \gamma)k_{f_j}(x,x; \theta)^{-1} y \\
\Cov(f^*)_j &= k_{f_j}(x^*, x^*; \gamma) - k_{f_j}(x^*, x; \gamma) k_{f_j}(x,x;\theta)^{-1}k_{f_j}(x,x^*;\gamma),
\end{aligned}
\end{equation}
where $k_{f_j}$ is the kernel associated with sample $g_j$ from the posterior over $g$ after transformation to a spectral density and then evaluation of the trapezoidal approximation (suppressing dependence on $\omega$ used in Eq.~\eqref{eqn: trap}).
$y$ is an $n \times 1$ vector of training data. $k_{f_j}(x,x; \theta)$ is an $n \times n$ matrix formed by evaluating $k_{f_j}$ at all pairs of $n$ training inputs $x$. Similarly $k_{f_j}(x^*, x^*; \theta)$ is a scalar and $k_{f_j}(x^*,x)$ is $1 \times n$ for a single test input $x^*$.
This distribution is a mixture of Gaussians with $J$ components. Following the above procedure, we obtain $J$ samples from the unconditional distribution in Eq.~\eqref{eq:pred}. We can compute the sample mean for point predictions and twice the sample standard deviation for a credible set. Alternatively, we can use the mixture of Gaussians representation in conjunction with the laws of total mean and variance to approximate the moments of the predictive distribution in Eq.~\eqref{eq:pred}, which is what we do for the experiments.

\section{Experiments}\label{sec: experiments}

We demonstrate the practicality of FKL over a wide range of experiments:
(1) recovering known kernels from data (Section \ref{sec: kernel-recovery}); (2) extrapolation (Section \ref{sec: extrapolation}); (3) multi-dimensional inputs and irregularly spaced data (section \ref{sec: multi-dim}); (4) multi-task precipitation data (Section \ref{sec: precip}); and (5) multidimensional pattern extrapolation (Section \ref{sec:texture}). We compare to the standard RBF and Mat\'ern kernels, as well as spectral mixture kernels \citep{wilson2013gaussian}, and the Bayesian nonparametric spectral estimation (BNSE) of \citet{tobar2018bayesian}.

For FKL experiments, we use $g(\omega)$ with a negative quadratic mean function (to induce an RBF-like prior mean in the distribution over kernels), and a Mat\'ern kernel with $\nu = \frac{3}{2}$ (to capture the typical sharpness of spectral densities). We use the heuristic for frequencies in the trapezoid rule described in Section \ref{spectral}. Using $J=10$ samples from the posterior over kernels, we evaluate the sample mean and twice the sample standard deviation from the unconditional predictive distribution in Eq.~\eqref{eq:pred} for point predictions and credible sets. We perform all experiments in GPyTorch \citep{gpytorch}.

\subsection{Recovery of Spectral Mixture Kernels}\label{sec: kernel-recovery}

\begin{figure}
  \centering
  \includegraphics[width=\linewidth]{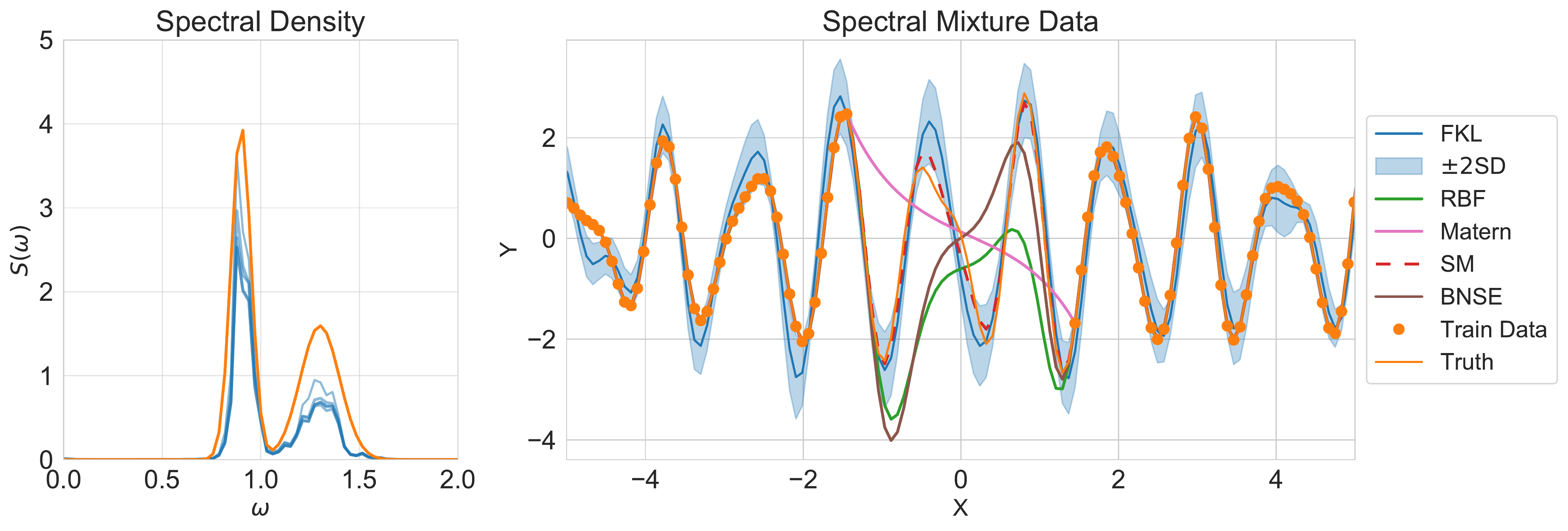}

  \caption{\textbf{Left}: Samples from the FKL posterior over the spectral density capture the shape of the true spectrum.
  \textbf{Right}: Many of the FKL predictions on the held out data are nearly on par with the ground-truth model (SM in dashed red). GPs using the other kernels perform poorly on extrapolation away from the training points.}

  \label{fig: sm-kernel}
\vspace{-0.5cm}
\end{figure}

Here we test the ability of FKL to recover known ground truth kernels. We generate 150 data points,  $x_i \sim U(-7.,7)$ randomly and then draw a random function from a GP with a two component spectral mixture kernel with weights 1 and 0.5, spectral means of 0.2 and 0.9 and standard deviations of 0.05.
As shown in Figure \ref{fig: sm-kernel}, FKL accurately reconstructs the underlying spectral density, which enables accurate in-filling of data in a held out test region, alongside reliable credible sets. A GP with a spectral mixture kernel is suited for this task and closely matches with withheld data. GP regression with the RBF or Mat\'ern kernels is unable to predict accurately very far from the training points. BNSE similarly interpolates the training data well but performs poorly on the extrapolation region away from the data. In Appendix \ref{app:qp} we illustrate an additional kernel recovery experiment, with similar results.

\subsection{Interpolation and Extrapolation}\label{sec: extrapolation}

\paragraph{Airline Passenger Data}
We next consider the airline passenger dataset \cite{hyndman_2005} consisting of 96 monthly observations of numbers of airline passengers from 1949 to 1961, and attempt to extrapolate the next 48 observations. We standardize the dataset to have zero mean and unit standard deviation before modeling. The dataset is difficult for Gaussian processes with standard stationary kernels, due to the rising trend, and difficulty in extrapolating quasi-periodic structure.
\paragraph{Sinc}
We model a pattern of three sinc functions replicating the experiment of \citet{wilson2013gaussian}. Here $y(x) = \text{sinc}(x + 10) + \text{sinc}(x) + \text{sinc}(x-10)$ with $\text{sinc}(x) = \sin(\pi x)/(\pi x)$.
This has been shown previously \citep{wilson2013gaussian} to be a case for which parametric kernels fail to pick up on the correct periodic structure of the data.

\begin{figure}[t!]
  \centering
  \begin{subfigure}[t]{.414\textwidth}
    \centering
    \includegraphics[width=\textwidth]{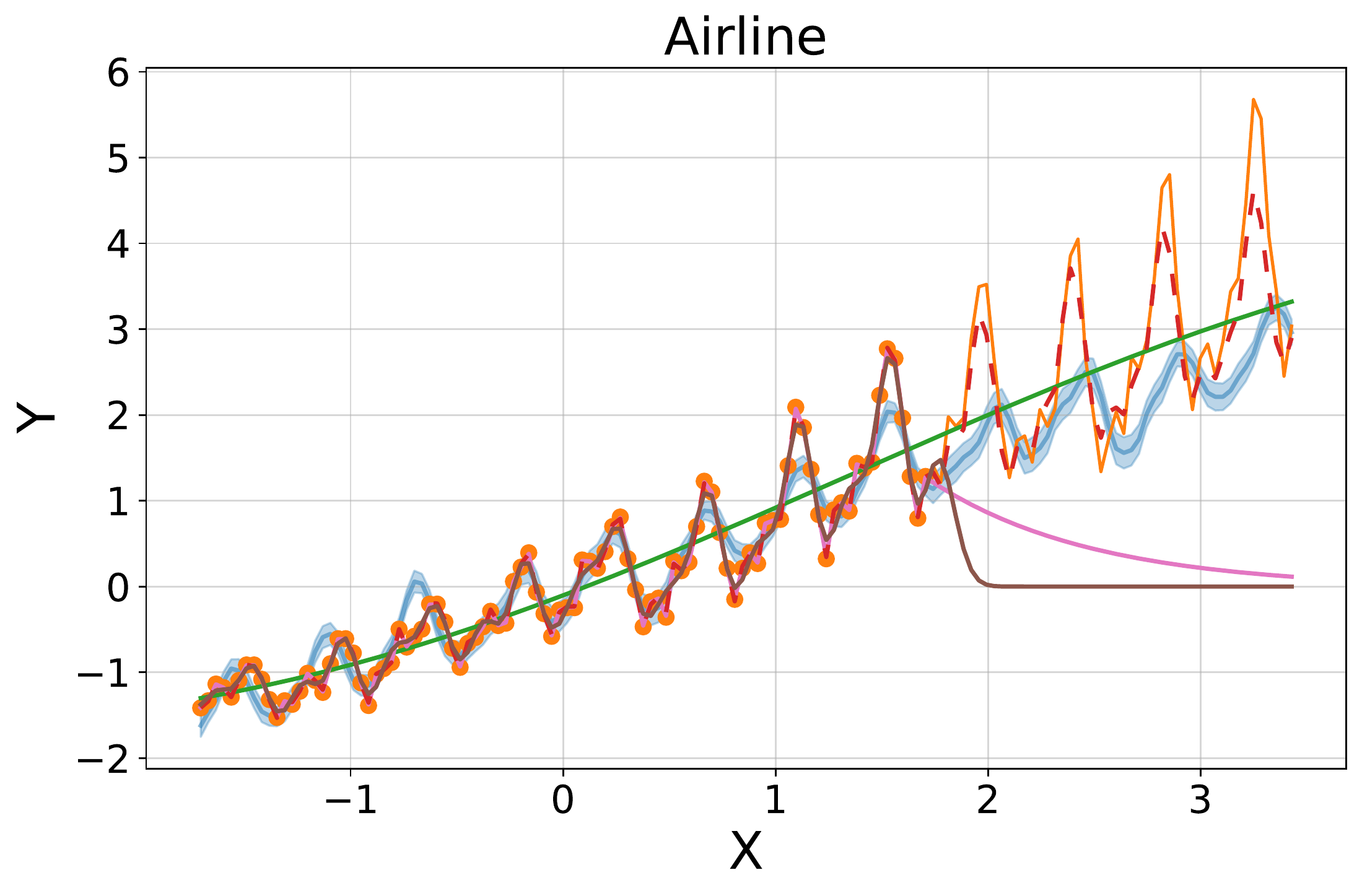}
    \caption{Extrapolation on the airlines dataset \cite{hyndman_2005}.}
    \label{fig:airline}
  \end{subfigure}
  \begin{subfigure}[t]{.536\textwidth}
  \centering
    \includegraphics[width=\textwidth]{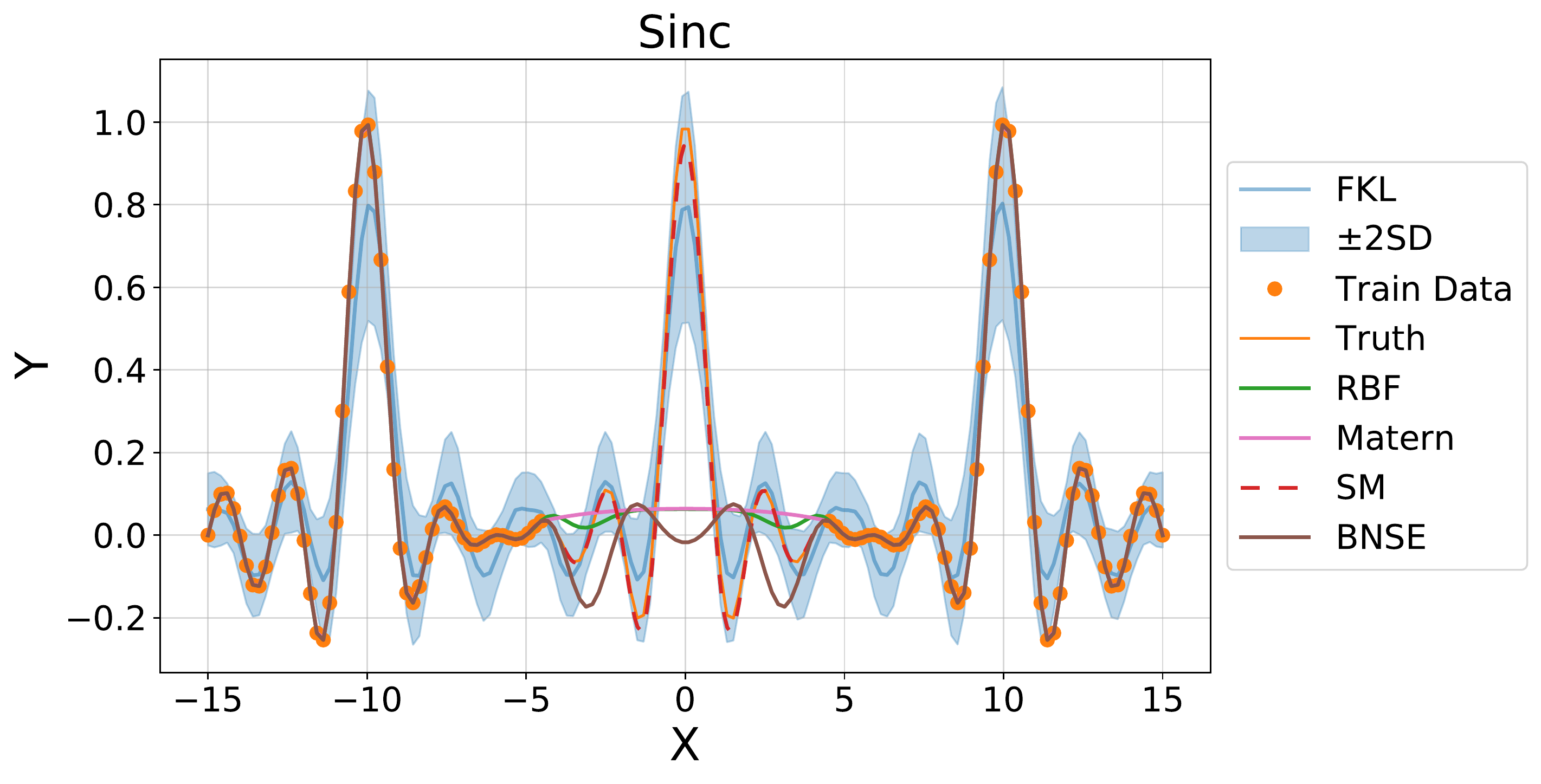}
  \caption{Interpolation on the $\text{sinc}$ function.}
    \label{fig:sinc}
  \end{subfigure}
  \caption{\textbf{(a)}: Extrapolation on the airline passenger dataset. \textbf{(b)}: Prediction on sinc data. FKL is on par with a carefully tuned SM kernel (dashed pink) in \textbf{(a)} and shows best performance in \textbf{(b)}, BNSE (brown) performs well on the training data, but quickly reverts to the mean in the testing set.}
\end{figure}

Figures \ref{fig:airline} and \ref{fig:sinc} show that FKL outperforms simple parametric kernels on complex datasets. Performance of FKL is on par with that of SM kernels while requiring less manual tuning and being more robust to initialization.

\subsection{Multiple Dimensions: Interpolation on UCI datasets}\label{sec: multi-dim}

\begin{figure}
	\centering
	\includegraphics[width=\textwidth]{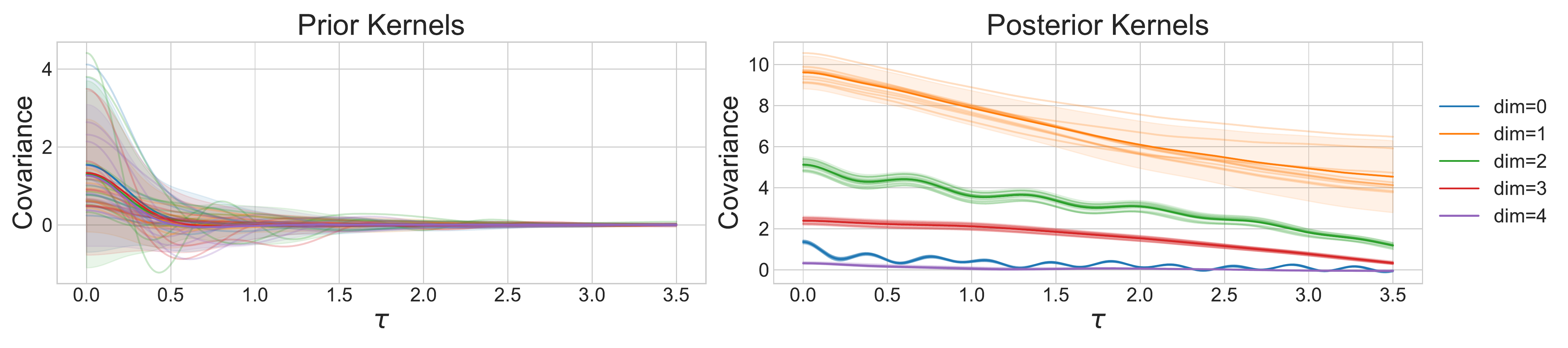}
	\caption{Samples of prior \textbf{(a)} and posterior \textbf{(b)} kernels displayed alongside the sample mean (thick lines) and $\pm$ 2 standard deviations (shade). Each color corresponds to a kernel, $k(\cdot)$, for a dimension of the airfoil dataset.}
	\label{fig:airfoil-prior-posterior-kernels}
\end{figure}

We use the product kernel described in Section \ref{sec: multi-dim} with both separate and shared latent GPs for regression tasks on UCI datasets. Figure \ref{fig:airfoil-prior-posterior-kernels} visually depicts the model with respect to prior and posterior products of kernels. We standardize the data to zero mean and unit variance and randomly split the training and test sets, corresponding to 90\% and 10\% of the full data, respectively. We conduct experiments over 10 random splits and show the average RMSE and standard deviation.
 We compare to the RBF, ARD, and ARD Mat\'ern. Furthermore, we compare the results of sharing a single latent GP across the kernels of the product decomposition(Eq. \ref{eq:multi_in_model}) with independent latent GPs for each kernel in the decomposition.

\begin{figure}
	\centering
	\includegraphics[width=\linewidth]{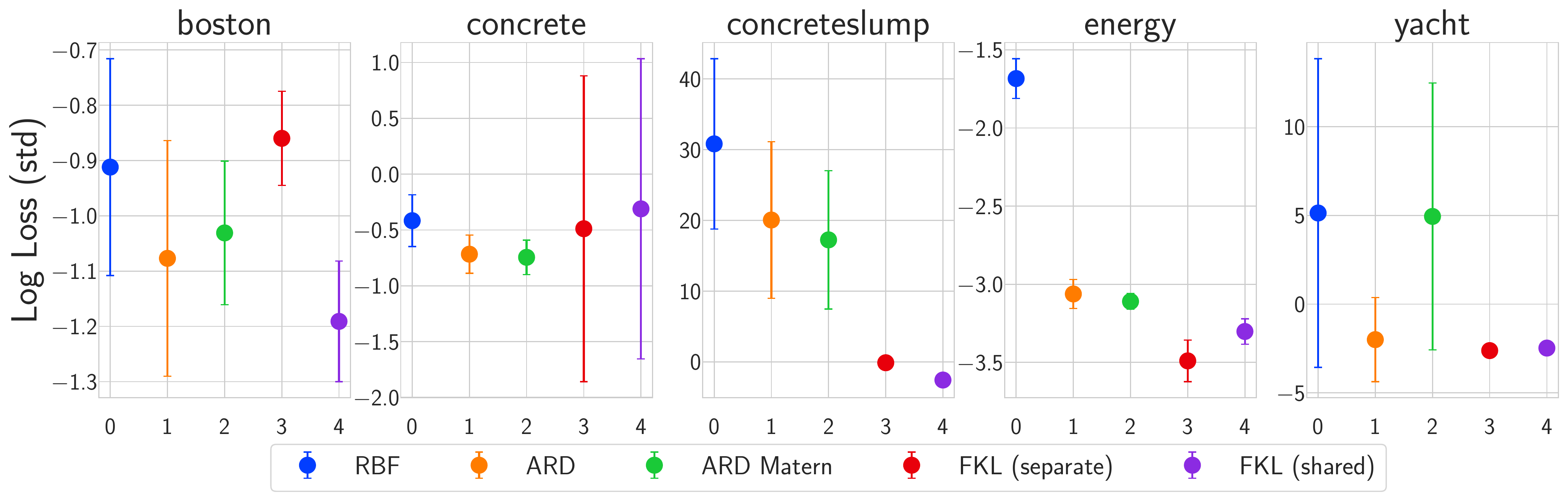}
	\caption{Standardized log losses on five of the 12 UCI datasets used. Here, we can see that FKL typically outperforms parametric kernels, even with a shared latent GP. See Table \ref{uc-table} for the full results in the Appendix.}
	\label{fig:uci}
\end{figure}

\subsection{Multi-Task Extrapolation}\label{sec: precip}

\begin{figure}[!htb]
  \includegraphics[width=\textwidth]{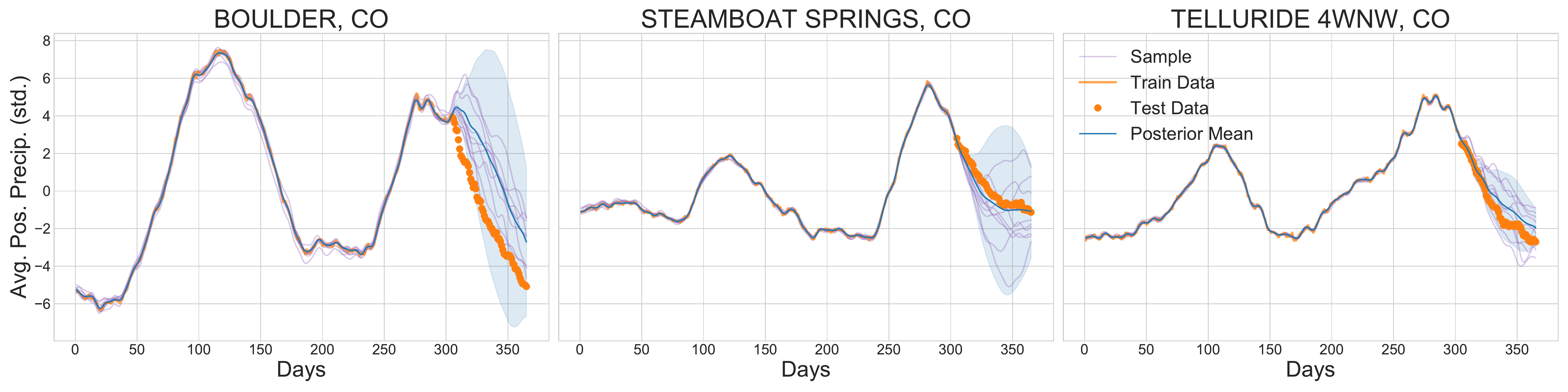}
  \caption{Posterior predictions generated using latent GP samples. 10 samples of the latent GP for each site are used to construct covariance matrices and posterior predictions of the GPs over the data.}
  \label{fig: prcp}
  \vspace{-0.5cm}
\end{figure}

We use the multi-task version of FKL in Section \ref{sec: multi-task} to model
precipitation data sourced from the United States Historical Climatology Network \cite{ushcn}.
The data contain daily precipitation measurements over 115 years collected at 1218 locations in the US. Average positive precipitation by day of the year is taken for three climatologically similar recording locations in Colorado: Boulder, Telluride, and Steamboat Springs, as shown in Figure \ref{fig: prcp}. The data for these locations have similar seasonal variations,
motivating a shared latent GP across tasks, with a flexible kernel process capable of learning this structure.
Following the procedure outlined in Section \ref{sec: inference} and detailed in Algorithm \ref{alg: mt-sampling} in the Appendix, FKL provides predictive distributions that accurately interpolates and extrapolates the data with appropriate credible sets.
In Appendix \ref{sec: big-prcp} we extend these multi-task precipitation results to large scale experimentation with datasets containing tens of thousands of points.

\subsection{Scalability and Texture Extrapolation}\label{sec:texture}

Large datasets typically provide additional information to learn rich covariance structure.
Following the setup in \cite{wilson2014fast}, we exploit the underlying structure in images and scale FKL to learn such a rich covariance --- enabling extrapolation on textures.
When the inputs, $X$, form a Cartesian product multidimensional grid, the covariance matrix decomposes as the Kronecker product of the covariance matrices over each input dimension, i.e. $K(X, X) = K(X_1, X_1) \otimes K(X_2, X_2) \otimes \cdots \otimes K(X_P, X_P)$ where $X_i$ are the elements of the grid in the $i^{th}$ dimension \cite{saatcci2012scalable}. Using the eigendecompositions of Kronecker matrices, solutions to linear systems and log determinants of covariance matrices that have Kronecker structure can be computed exactly in $\mathcal{O}(P N^{P/2})$ time, instead of the standard cubic scaling in $N$ \citep{wilson2014fast}.

\begin{figure}
\centering
\includegraphics[width=\linewidth]{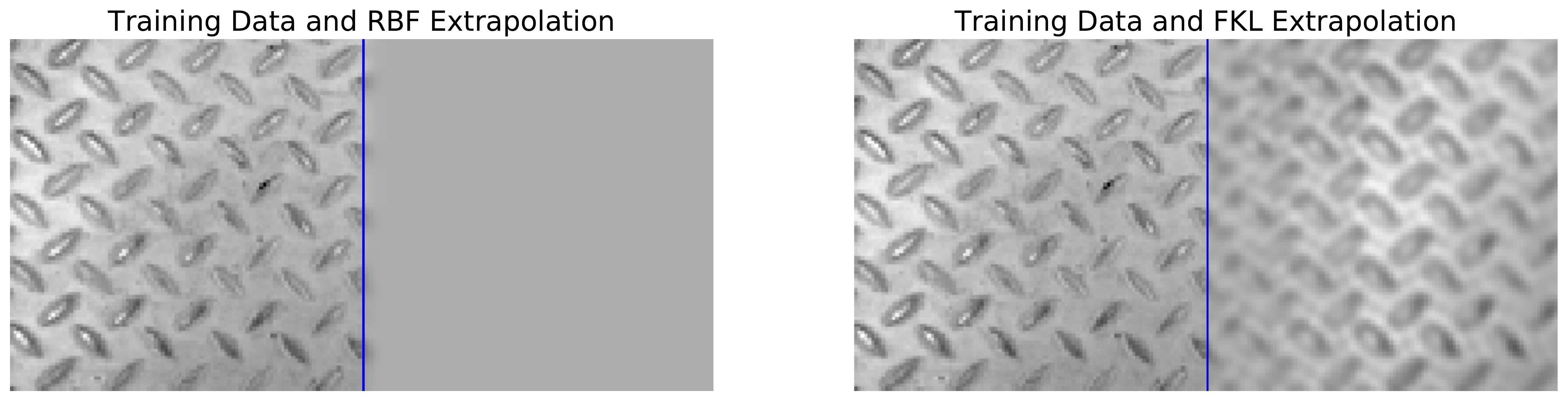}
\caption{Texture Extrapolation: training data is shown to the left of the blue line and predicted extrapolations according to each model are to the right.}
\label{fig:texture}
\end{figure}

We train FKL on a $10,000$ pixel image of a steel tread-plate and extrapolate the pattern beyond the training domain. As shown in Figure \ref{fig:texture}, FKL uncovers the underlying structure, with no sophisticated initialization procedure. While the spectral mixture kernel performs well on these tasks \citep{wilson2014fast}, it requires involved initialization procedures. By contrast, standard kernels, such as the RBF kernel, are unable to discover the covariance structure to extrapolate on these tasks.

\section{Discussion}

In an era where the number of model parameters often exceeds the
number of available data points, the function-space view provides a more natural representation of our models. It is the complexity and inductive biases of \emph{functions} that affect generalization performance, rather than the number of parameters in a model. Moreover, we can interpretably encode our assumptions over functions, whereas parameters are often inscrutable.
We have shown the function-space approach to learning covariance structure is flexible and convenient, able to automatically discover rich representations of data, without over-fitting.

There are many exciting directions for future work: (i) interpreting the learned covariance structure across multiple heterogeneous tasks to gain new scientific insights; (ii) developing function-space distributions over \emph{non-stationary} kernels; and (iii) developing deep hierarchical functional kernel learning models, where we consider function space \emph{distributions over distributions} of kernels.

\section*{Acknowledgements}
GWB,  WJM,  JPS,  and  AGW  were  supported  by  an  Amazon Research Award, Facebook Research, NSF IIS-1563887, and NSF IIS-1910266.  WJM was additionally supported by an NSF Graduate Research Fellowship under Grant No.  DGE-1650441.

\bibliographystyle{plainnat}
\bibliography{refs}

\cleardoublepage
\appendix

\begin{algorithm}[tb]
	\caption{Alternating Sampler}
	\label{alg: alt-sampler}
	\begin{algorithmic}
		\STATE {\bfseries Input:} Data $(x, y)$, Initial hyper-parameters $\phi_{0}$,
		Sampling frequencies $\omega$, Initial Latent GP $g(\omega)$,
		Number of gradient steps to take per iteration $N_{optim}$,
		Number of ESS samples per update per iteration $N_{ESS}$,

		\REPEAT

		\FOR{$i=1$ {\bfseries to} $N_{optim}$}
		\STATE Update $\phi$ using gradient descent given $g(\omega)$ and Eqn. \ref{eqn: theta-loss}
		\ENDFOR
		\FOR{$i=1$ {\bfseries to} $N_{ESS}$}
		\STATE Update $g(\omega)$ using elliptical slice sampling given $\phi$ and Eqn. \ref{eqn: g-loss}
		\ENDFOR
		\UNTIL{convergence}
	\end{algorithmic}
\end{algorithm}

\section{Computational Complexity}\label{sec: scalability}
Note that when sampling at $N$ data points and $I$ frequencies, the storage costs for this model are naively $\mathcal{O}(N^2 + I^2)$ with the computational cost for prediction of $\mathcal{O}(N^3 + I^3).$
Using pre-conditioned conjugate gradients for inverses and stochastic Lanczos quadrature (SLQ) or the log determinants \citep{dong_scalable_2017} as implemented in GPyTorch \citep{gpytorch} for the data and likelihood calls can immediately reduce the computational cost to $\mathcal{O}(N^2 + I^2).$
However, the randomness in the log determinant calculations proved to be problematic for ESS and we only used SLQ for the gradient-based updates, keeping the overall time complexity cubic in $N$.
Given that the latent Gaussian processes are on a pre-defined grid, we can utilize fast Toeplitz matrix multiplications \citep{wilson2014fast} to reduce the time complexity to $\mathcal{O}(N^3 + I \log{I})$ and the memory complexity to $\mathcal{O}(N^3 + I).$

Extending the model to multi-dimensional inputs and multiple outputs adds on a linear term for both dimensionality $D$ and tasks $T$ independently, so for a multi-task model with $T$ tasks predictions are done in $\mathcal{O}(T(N^3 + I))$. Note that this is significant improvement over the $\mathcal{O}(T^3 N^3)$ needed to do exact inference in previous multi-task work such as \citet{bonilla2008multi}.

For enhanced scalability, we can approximate the kernel matrices in single (and low) dimensions by utilizing scalable kernel interpolation (SKI) as introduced by \citet{wilson2015kernel}. Using $m$ inducing points we can achieve an inference cost of $\mathcal{O}(N + m \log m + I \log I)$ or $\mathcal{O}(T(N + m \log m + I \log I))$ for the multi-task setting.

\section{Latent Model Specification}\label{app:lm}
\subsection{Initialization}
FKL proves to be robust to initialization, thus for simplicity we initialize the spectral density to be constant, $S(\omega) = 1$, for a large range of frequencies. An experiment detailing the models robustness is given in the Appendix.

\subsection{Specification of the Latent GP}
We fix the mean and covariance of the latent process $g(\omega)$ to take the following forms:
\begin{equation}\label{eqn: prior-setup}
\begin{aligned}
&\{\text{log of RBF spectral density}\} \hspace{2cm} \mu(\omega;\theta) = \theta_0 - \frac{\omega^2}{2\tilde{\theta_1}^2}\\
&\{\text{Mat\'ern kernel}\}\hspace{0.8cm}k_g(\omega, \omega';\theta) = \frac{2^{1-\nu}}{\Gamma(\nu)}  \left(\sqrt{2\nu}\frac{|\omega-\omega'|}{\tilde{\theta_2}}\right)K_{\nu}\left(\sqrt{2\nu}\frac{|\omega-\omega'|}{\tilde{\theta_2}}\right) + \tilde\theta_3\delta_{\tau=0}
\end{aligned}
\end{equation}
The $\tilde\theta_i$'s are non-negative variables, so are computed with $\tilde\theta_i = \log(e^{\theta_i} + 1)$, the softplus of the raw value.
The mean parametrization coupled with the constraints fixes the latent mean to be negative quadratic, like the logarithm of an RBF spectral density.

\subsection{Prior Specification}\label{sec: hyperprior}
For the noise terms, we place smoothed box priors\footnote{A smooth approximation to uniform priors, where $B(x) = \{a \leq x \leq b\}$ then $d(x, B):= \min_{x' \in B}|x - x'|$ and finally the density is given by $f(x):= \exp\{-d(x,B)^2 / \sqrt{2 \sigma^2}\}.$ See \url{https://gpytorch.readthedocs.io/en/latest/priors.html} for further implementation details.} on the range (1e-8, 1e-3) to control both numerical instability and the noise terms.
For the constant mean terms in both the data and latent means, we place uninformative $\mathcal{N}(0, 100)$ priors.
For the length-scale in the spectral density mean along with the length-scale and output-scale of the covariance of the spectral density GP, we place standard log-normal priors.

\begin{algorithm}[tb]
	\caption{Multi-Task Alternating Sampler}
	\label{alg: mt-sampling}
	\begin{algorithmic}
		\STATE {\bfseries Input:} Data $(x, Y)$, Initial hyper-parameters $\phi_{0}$,
		Sampling frequencies $\omega$, Initial Latent GPs $g_i(\omega)$ for $i=1,\dots, T$,
		Number of gradient steps to take per iteration $N_{optim}$,
		Number of ESS samples per update per iteration $N_{ESS}$,

		\REPEAT

		\FOR{$i=1$ {\bfseries to} $N_{optim}$}
		\STATE Update $\phi$ using gradient descent given $g(\omega)$ and Eqn. \ref{eqn: mt-theta-loss}
		\ENDFOR
		\FOR{$t=1$ {\bfseries to} $T$}
		\FOR{$i=1$ {\bfseries to} $N_{ESS}$}
		\STATE Update $g_t(\omega)$ using elliptical slice sampling given $\phi$ and Eqn. \ref{eqn: g-loss} with respect to $f_t(x)$
		\ENDFOR
		\ENDFOR
		\UNTIL{convergence}
	\end{algorithmic}
\end{algorithm}

\section{Density and Error Bounds of FKL}\label{app:theory}

\subsection{Error Rate of Trapezoidal Rule Approximation}

Given a sample path from a Gaussian process with a Mat\'ern kernel as is used in our implementation, we can get explicit $O(1/I)$ error bounds on the error of trapezoid rule integration of the warped GP instead by checking Holder continuity of sample draws from the latent GP \citep{belyaev1961continuity}, and using results on the error of trapezoid rule for Holder continuous functions \citep{cruz2002sharp}. Note that we could use standard error bounds if we use a GP with twice differentiable sample paths.

\subsection{Density Amongst Stationary Kernels}

We next note that the trapezoidal rule is just a finite sample version of both Riemann and Darboux integrals.
Thus, functional kernel learning can also be written as a linear combination of the trigonemetric basis expansions and the spectral density (e.g. in sparse spectrum form like \citet{lazaro-gredilla_sparse_nodate}).
Thus, FKL can model discontinuous but finite measures because mixtures of Gaussians are dense approximations of Riemann integrable densities (see Theorem 5 of \citet{shen2019harmonizable}).
Thus, the trapezoid rule will be an approximator of the true kernel on the compact set $[0, \omega_{max}]$, converging as $\omega_{max} \rightarrow \infty$ (e.g. as the number of basis functions goes to infinity).

Finally, we note that in the multi-dimensional case, FKL does not provide support over all stationary covariances (like other spectral approaches \citep{shen2019harmonizable,wilson2013gaussian}), but we find in practice that the domain of support is great enough for accurate performance on most tasks.
We would need to at least model the $\omega$'s for each dimension on a grid to provide full support, at a cost of af the number of grid points exponentially increasing.
Future work will help to alleviate this issue.

\section{Sensitivity to Initialization}\label{app:sensitivity}

Part of the strength of FKL, particularly over competing methods like spectral mixture (SM) kernels, is robustness to initialization. We compare the performance of FKL and SM kernels on interpolating data generated from a GP with a quasi-periodic kernel.

In GPyTorch spectral mixture (SM) kernels are initialized to,
\begin{align*}
\mu &= \log(\exp(0) + 1)\\
\sigma &= \log(\exp(0) + 1)\\
w &= \log(\exp(0) + 1),
\end{align*}
i.e. the means, variances, and weights of each mixture component is the softplus of 0 prior to calling the data initialization routine \cite{gpytorch}. The data-based initialization routine uses statistics of the data to randomly initialize the parameters of the mixture components, and performance is highly dependent on this initialization.

In the current implementation FKL is initialized with a spectral density that is constant,
\begin{align*}
S_0(\omega) &= 1 \quad \forall \omega\\
g_0(\omega) &= 0 \quad \forall \omega,
\end{align*}
where $g(\omega)$ is the log-spectral density, which is modeled using a latent GP.
The surprising fact, and what makes FKL such an appealing model for complex problems, is robustness to initialization. In practice we see no gains in predictive performance when initializing in a more sophisticated fashion than is currently done. This robustness goes far enough that we don't even see performance gains when we have access to ground truth data and can initialize the spectral density to be near to the spectral density of the kernel of generative model itself.

Data are generated using a GP with quasi-periodic kernel and the middle portion of the data are held out as a testing set. Using the inverse Fourier transform we can compute the spectral density of the generating quasi-periodic kernel directly, $S^*(\omega)$.
First we train and predict using a SM kernel that is has parameters initialized to the constant values from above, and compare to a SM kernel using GPyTorch's built in data-based initialization.
Next we repeat the procedure using a default initialized FKL model, then compare to an FKL model where the spectral density has been initialized to a corrupted version of the ground truth spectral density. Thus we compare FKL models with the initializations,
\begin{align*}
S_0(\omega) &= 1 \quad \forall \omega \\
S_0(\omega) &= S^*(\omega) + \mathcal{N}(0, 0.1) \quad \forall \omega.
\end{align*}

\begin{figure}
	\begin{subfigure}{0.5\linewidth}
		\centering
		\includegraphics[width=\linewidth]{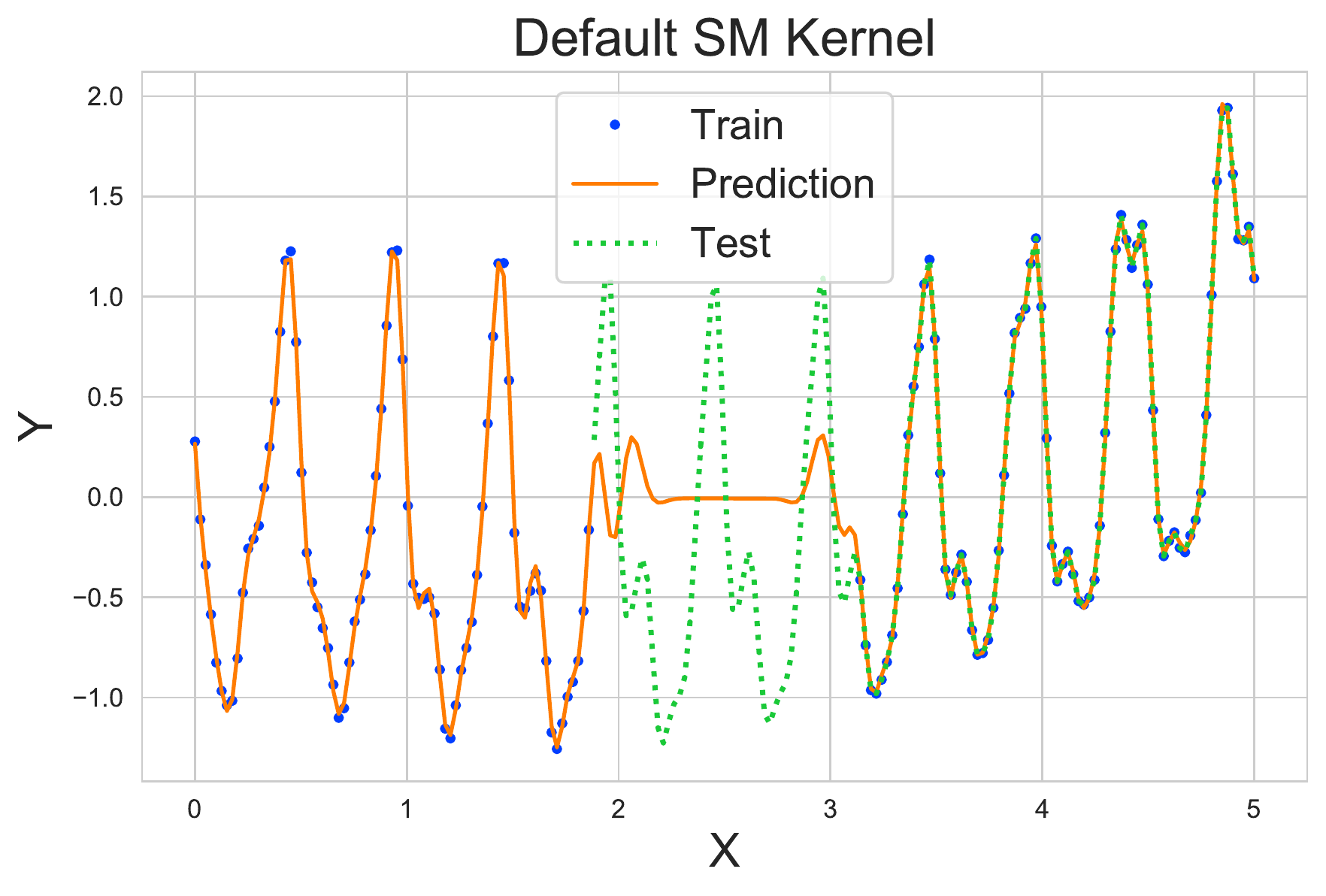}
	\end{subfigure}
	\begin{subfigure}{0.5\linewidth}
		\centering
		\includegraphics[width=\linewidth]{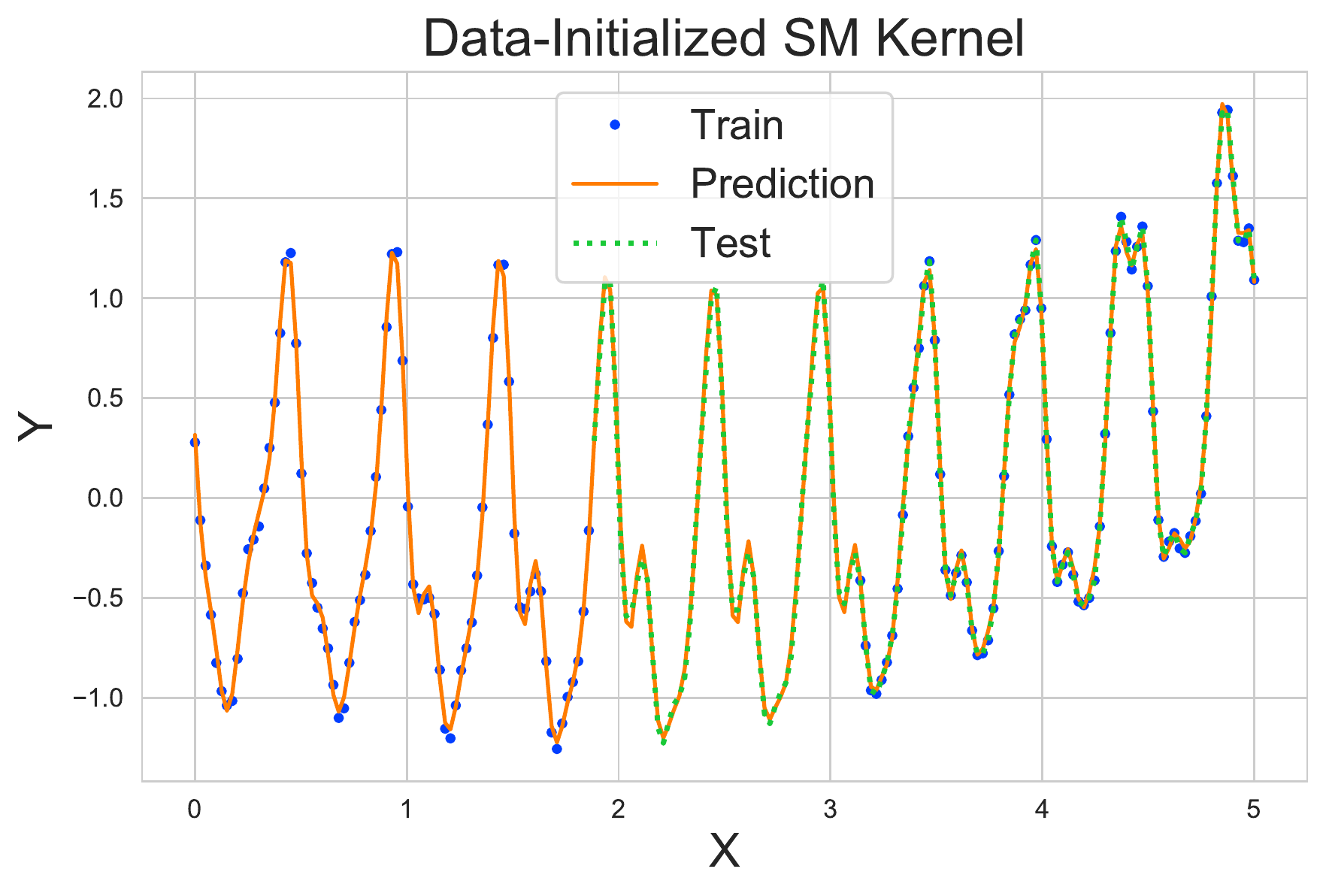}
	\end{subfigure}\\[1ex]
	\caption{Comparison of naive and data-based initialized SM kernels on interpolation tasks. \textbf{Left}: the default (naive) initialized kernel, \textbf{Right}: the data-based initialized kernel.}
	\label{fig: sm-init}
\end{figure}

\begin{figure}
	\begin{subfigure}{0.5\linewidth}
		\centering
		\includegraphics[width=\linewidth]{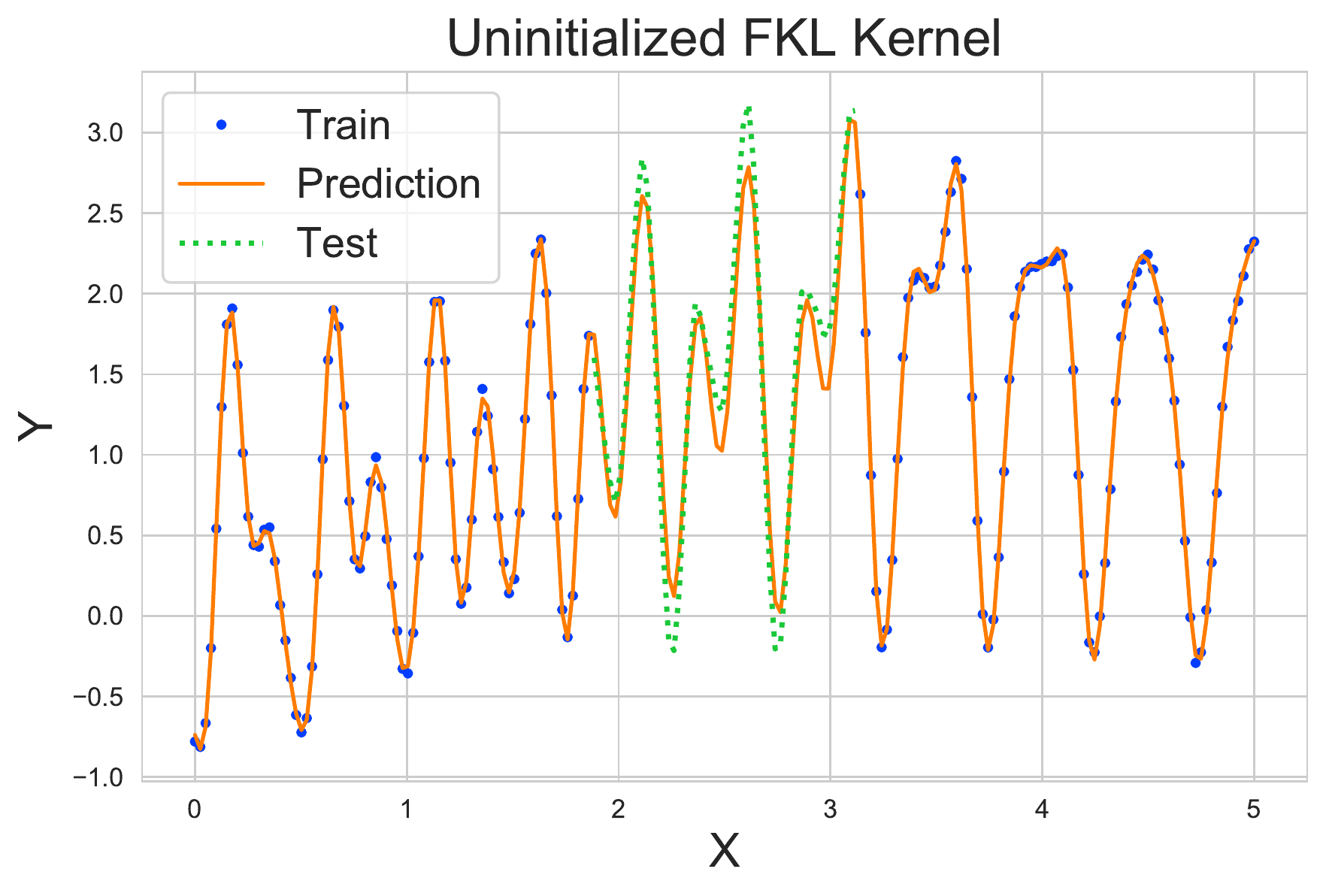}
	\end{subfigure}
	\begin{subfigure}{0.5\linewidth}
		\centering
		\includegraphics[width=\linewidth]{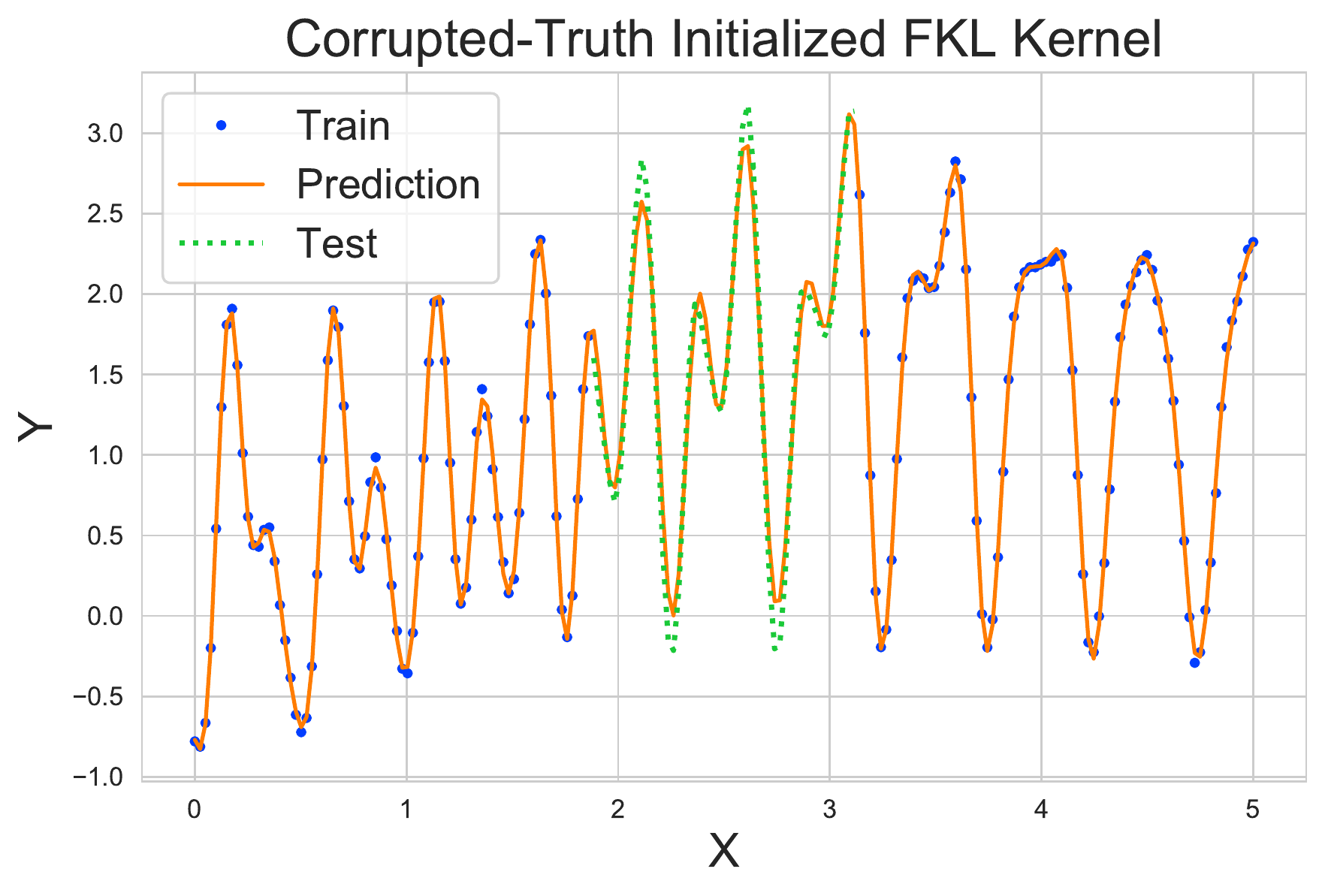}
	\end{subfigure}\\[1ex]
	\caption{Comparison of basic and ground-truth initialized FKL kernels on interpolation tasks. \textbf{Left}: the default (naive) initialized kernel, \textbf{Right}: the ground-truth initialized kernel.}
	\label{fig: fkl-init}
\end{figure}

The results are shown in Figures \ref{fig: sm-init} and \ref{fig: fkl-init}. What we see is that a naive implementation of SM kernels leads to poor performance on the testing set, while FKL performs nearly the same whether we initialize the spectral density to an arbitrary value, or to nearly the ground truth.

\section{Further Experiments}
\subsection{Recovery of Known Kernels}\label{app:qp}
\paragraph{Spectral Mixture Kernel}

\begin{figure}
	\begin{subfigure}{.5\linewidth}
		\centering
		\includegraphics[width=\linewidth]{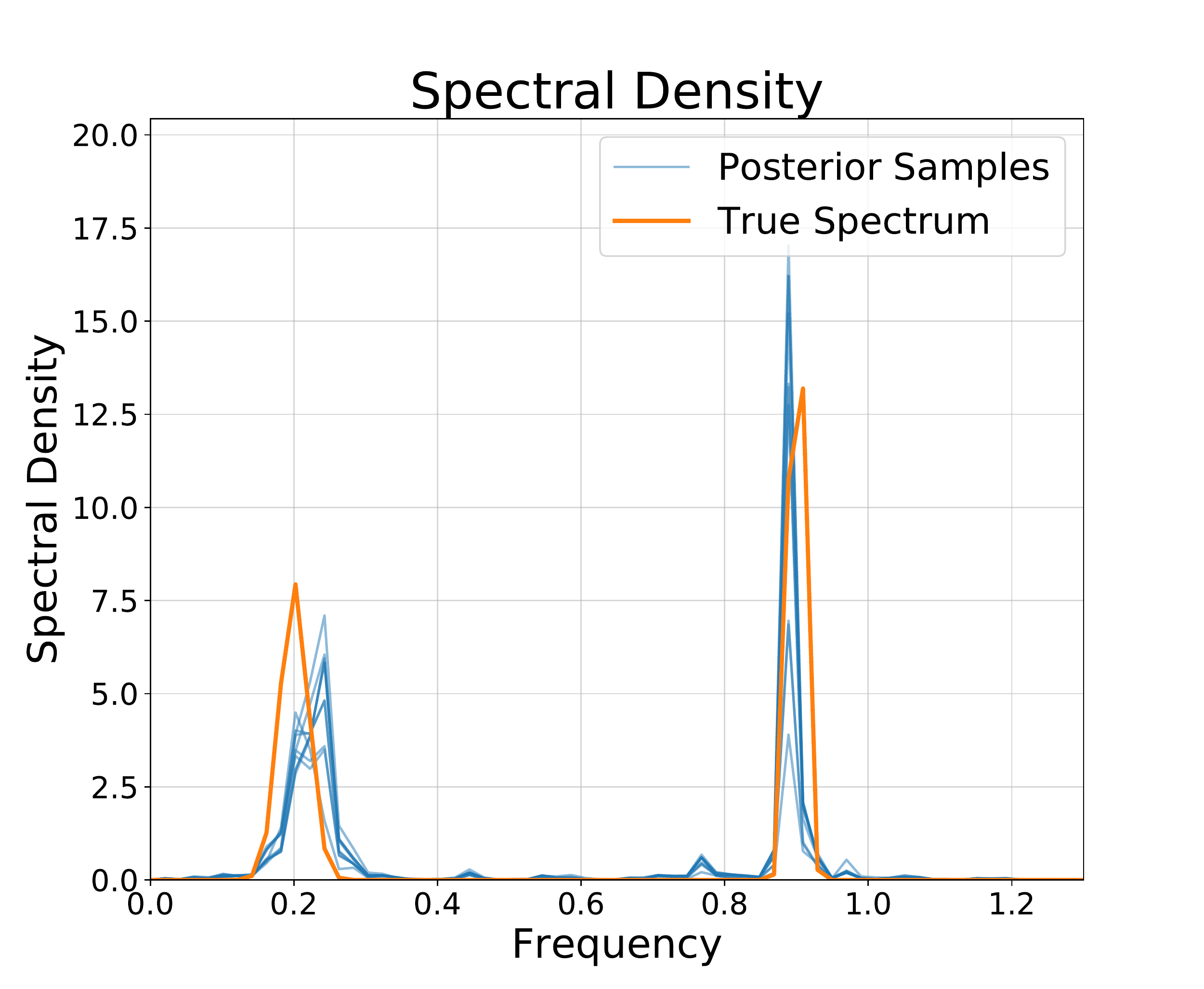}
	\end{subfigure}
	\begin{subfigure}{.5\linewidth}
		\centering
		\includegraphics[width=\linewidth]{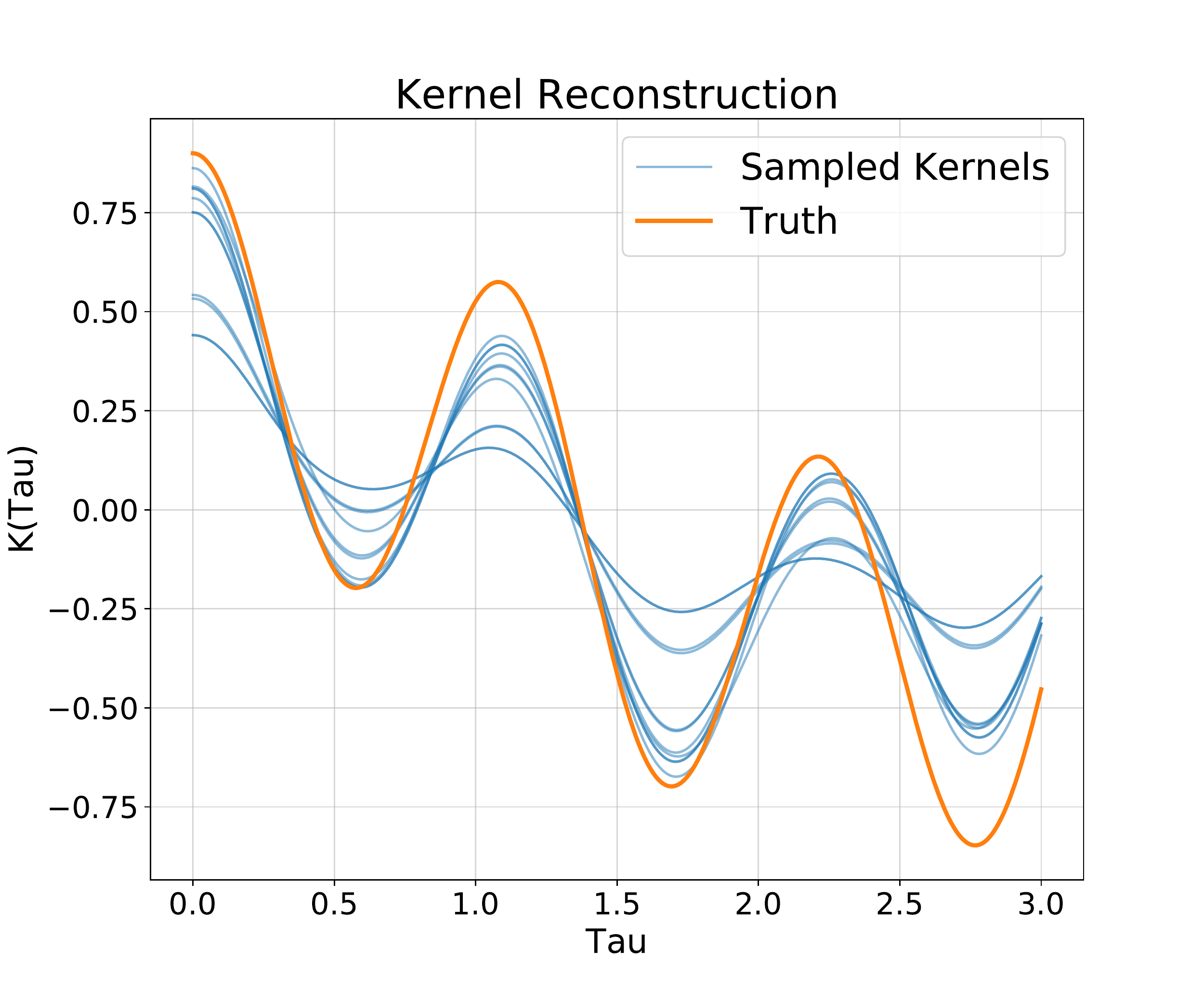}
	\end{subfigure}\\[1ex]
	\caption{Samples from the latent GP displayed in the spectral domain along with the ground truth (Left) and the reconstructed kernels generated by these samples (Right).}
	\label{fig: sm-appendix}
\end{figure}

Extending from Section \ref{sec: kernel-recovery}, we also display the accuracy of the kernel reconstruction given the samples drawn in the latent space. Figure \ref{fig: sm-appendix} shows the accurately sampled spectral density, and the kernels reconstructed from these samples.

\paragraph{Quasi-Periodic Kernel} Synthetic data are generated from a mean zero Gaussian process with kernel,
\begin{equation}
k(\tau; \ell, \omega) = \exp\left(-\frac{\tau^2}{2\ell^2}\right) \exp \left( - 2 \sin^2(\pi \tau \omega)\right).
\end{equation}
Since there is inherent periodicity in the generative model, the true spectral density has distinct modes corresponding to the period length of the sinusoidal component of the kernel. The spectral density of this kernel is not analytically computed, however using the known kernel the discrete Fourier transform allows an approximation of the ground-truth spectrum to be found, and comparison in the spectral domain can be made.

Using this latent GP model accurate reconstruction of both the spectral density and kernel are obtained using only training data. Further more, infilling into the testing set shows high accuracy and the confidence region encompasses the data.

\begin{figure}
	\begin{subfigure}{.5\linewidth}
		\centering
		\includegraphics[width=\linewidth]{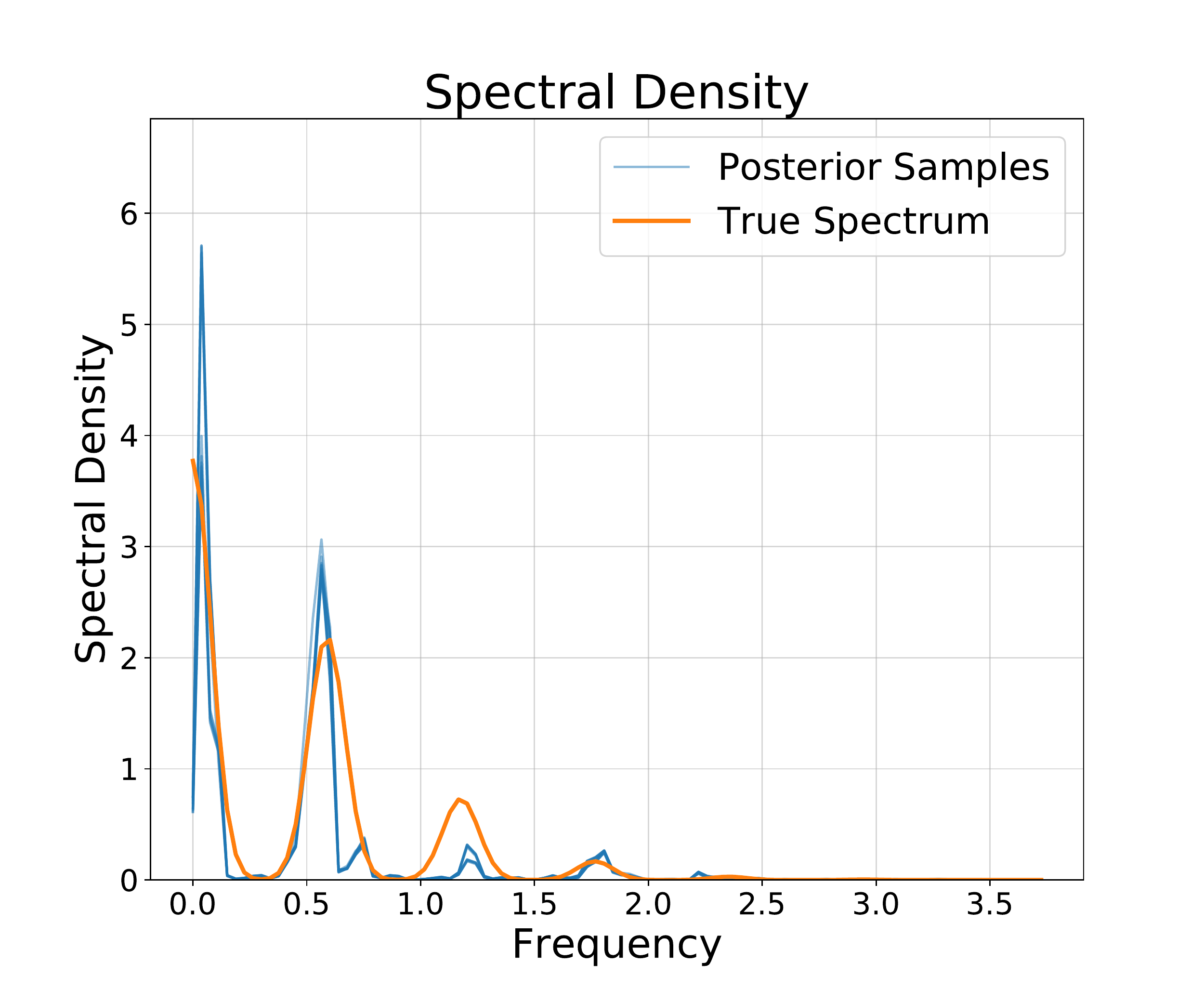}
	\end{subfigure}
	\begin{subfigure}{.5\linewidth}
		\centering
		\includegraphics[width=\linewidth]{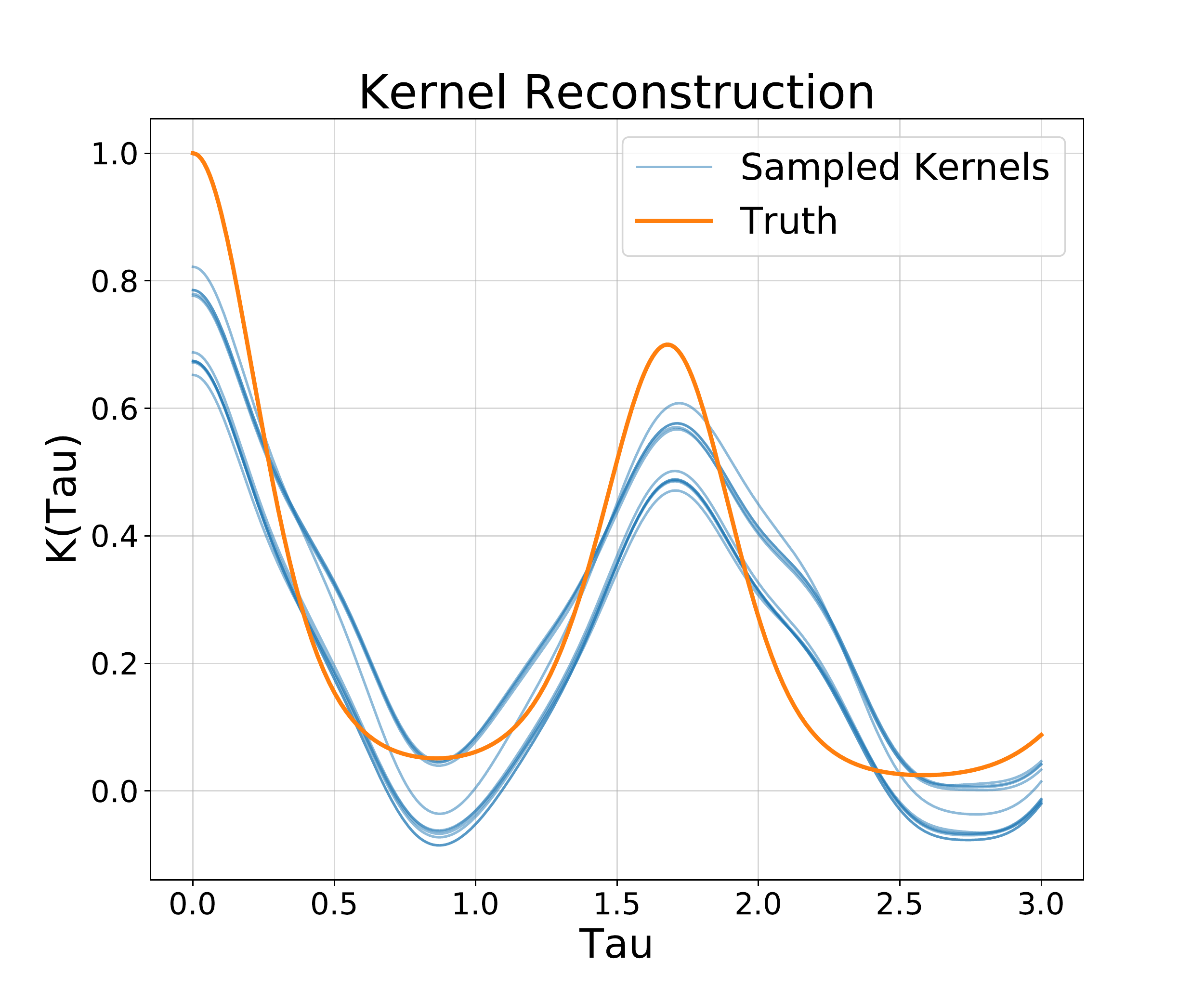}
	\end{subfigure}\\[1ex]
	\begin{subfigure}{\linewidth}
		\centering
		\includegraphics[width=\linewidth]{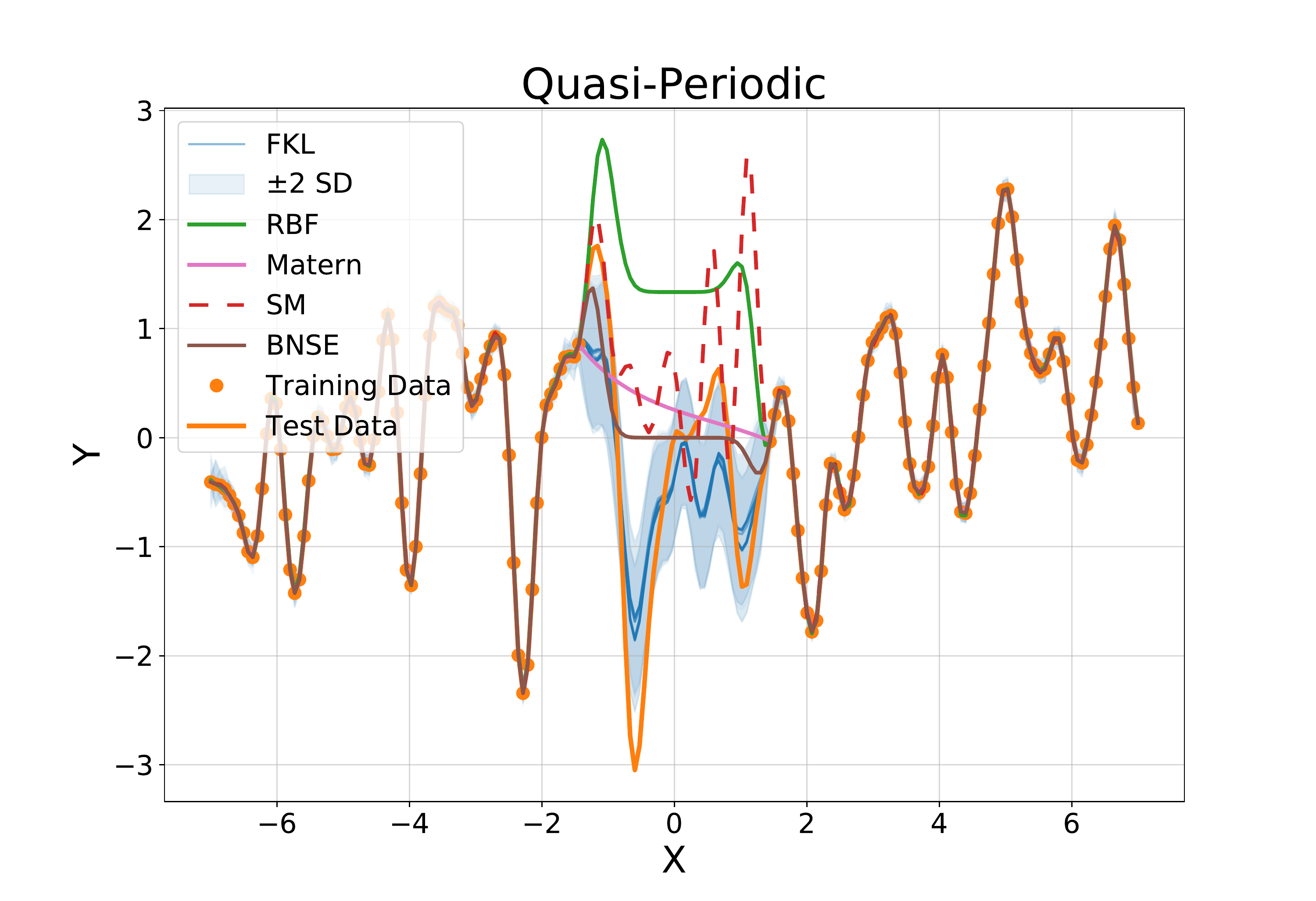}
	\end{subfigure}
	\caption{Spectrum (Above Left) and kernel (Above Right) reconstruction, and resulting data prediction (Below) for data generated by a quasi-periodic kernel.}
	\label{fig: qp-kernel}
\end{figure}

\subsection{Foreign Exchange Rates Dataset}
We consider multi-output prediction tasks on a foreign exchange rates dataset originally developed in \cite{alvarez2010efficient}.
The dataset consists of the exchange rates of 10 currencies and 3 precious metals with respect to the US dollar in 2007.
The task is to predict the Canadian dollar (CAD) on days 50-100, Japanese yen (JPY) on days 100-150, and Australian dollar (AUD) on days 150-200, given the exchange rate information for all other days.
Due to market differences, there are occasionally also missing data. Like in \cite{requeima2018gaussian}, we measure performance with the standardized mean square error (SMSE).
The results from this experiment are shown in Table \ref{tab:fx} with comparisons taken from both \cite{requeima2018gaussian} and \cite{nguyen2014collaborative}.
FKL performs considerably better than both types of collaborative Gaussian process, which constrain the outputs considerably more.
By comparison, the GPAR \citep{requeima2018gaussian} outperforms FKL on this task, perhaps due to its explicit ordering of tasks and its increased depth (the GPAR is a special case of deep Gaussian processes \citep{damianou2013deep}).

Here, we utilize 5 rounds of the alternating sampler with 10 optimization and 50 ESS iterations and run on a single GPU (with 10 repetitions taking about 3 minutes).

\begin{table}
	\centering
	\caption{Standardized mean squared error on FX dataset. Comparisons are with independent Gaussian processes (IGP), convolved multi-output GP (CMOGP) \citep{alvarez2011computationally}, collaborative GP (CGP) \citep{nguyen2014collaborative}, and Gaussian process autoregressive model (GPAR). Note that the GPAR is perhaps best viewed as a deep Gaussian process with known inputs. Comparisons taken from \citep{requeima2018gaussian}. Note that FKL multi-task outperforms the standard multi-task GP methods) averaged over 10 random trials.}
	\label{tab:fx}
	\begin{tabular}{lccccc}
		\toprule
		Model & IGP & CMOGP & CGP & GPAR & FKL(multi-task) \\
		\midrule
		SMSE & 0.5996 & 0.2427 & 0.2125 & \textbf{0.0302} & $0.1392 \pm 0.01$\\
		\bottomrule
	\end{tabular}
\end{table}

\subsection{UCI Tables}

Tables \ref{uc-table}, \ref{uc-table-msll}, and \ref{uc-table-nll} show the RMSE, standardized log loss, and negative log likelihoods of FKL (both separate and shared latent models) compared to standard parametric models on UCI regression tasks.

\begin{table}
	\caption{UCI Regression RMSEs, comparisons are with RBF, ARD, and ARD Mat\'ern kernels, $N$ points $D$ input dimensions. We compare to separate latent GPs for each input dimension, finding that sharing a single latent GP across dimensions works better than both the standard fixed spectrum approaches and separate latent GPs. Each of the experiments were conducted 10 times with random 90/10 train/test splits and we report the average RMSE $\pm$ one standard deviation.}
	\label{uc-table}
	\small
	\centering
	\resizebox{\textwidth}{!}{
		\begin{tabular}{cccccccc}
			\toprule
			Dataset & $N$ & $D$ & RBF & ARD & ARD Mat\'ern & FKL-PB (separate) & FKL-PB (shared) \\
			\midrule
			challenger & 23 & 4 & 0.713 $\pm$ 0.348 & 0.659 $\pm$ 0.368 & 0.612 $\pm$ 0.268 & 0.58 $\pm$ 0.225 & \textbf{0.548 $\pm$ 0.174} \\
			fertility & 100 & 9 & 0.159 $\pm$ 0.036 & 0.177 $\pm$ 0.035 & \textbf{0.148 $\pm$ 0.038} & 0.19 $\pm$ 0.047 & 0.182 $\pm$ 0.022 \\
			concreteslump & 103 & 7 & 36.302 $\pm$ 7.934 & 27.377 $\pm$ 7.782 & 26.335 $\pm$ 7.482 & 59.444 $\pm$ 12.879 & \textbf{4.385 $\pm$ 1.332} \\
			servo & 167 & 4 & 0.305 $\pm$ 0.056 & \textbf{0.23 $\pm$ 0.075} & 0.256 $\pm$ 0.06 & 0.282 $\pm$ 0.086 & 0.288 $\pm$ 0.063 \\
			yacht & 308 & 6 & 0.17 $\pm$ 0.07 & 0.187 $\pm$ 0.078 & 0.269 $\pm$ 0.048 & 0.193 $\pm$ 0.13 &\textbf{ 0.11 $\pm$ 0.054} \\
			autompg & 392 & 7 & 2.651 $\pm$ 0.488 & 3.077 $\pm$ 0.544 & \textbf{2.516 $\pm$ 0.332 }& 2.838 $\pm$ 0.374 & 2.69 $\pm$ 0.492 \\
			housing & 506 & 13 & 3.771 $\pm$ 0.675 & 3.222 $\pm$ 0.846 & 3.261 $\pm$ 0.624 & 4.679 $\pm$ 0.632 & \textbf{2.703 $\pm$ 0.227} \\
			stock & 536 & 11 & \textbf{0.005 $\pm$ 0.001} & \textbf{0.005 $\pm$ 0.001} & \textbf{0.005 $\pm$ 0.001} & 0.018 $\pm$ 0.002 & 0.016 $\pm$ 0.001 \\
			pendulum & 630 & 9 & 1.297 $\pm$ 0.315 & 1.185 $\pm$ 0.326 & \textbf{1.013 $\pm$ 0.207} & 2.747 $\pm$ 0.737 & 1.562 $\pm$ 0.554 \\
			energy & 768 & 8 & 1.839 $\pm$ 0.253 & 0.457 $\pm$ 0.035 & 0.373 $\pm$ 0.062 & \textbf{0.296 $\pm$ 0.066} & 0.334 $\pm$ 0.063 \\
			concrete & 1030 & 8 & 7.001 $\pm$ 0.513 & 6.125 $\pm$ 0.456 & 6.058 $\pm$ 0.373 & \textbf{3.781 $\pm$ 0.501} & 4.047 $\pm$ 0.693 \\
			airfoil & 1503 & 5 & 2.503 $\pm$ 0.202 & 1.696 $\pm$ 0.243 & 1.595 $\pm$ 0.296 & \textbf{1.378 $\pm$ 0.176 }& 1.39 $\pm$ 0.181 \\
			\bottomrule
		\end{tabular}
	}
\end{table}

\begin{table}
	\caption{UCI Regression Mean Standardized Log loss, comparisons are with RBF, ARD, and ARD Mat\'ern kernels, $N$ points $D$ input dimensions. We compare to separate latent GPs for each input dimension. Each of the experiments were conducted 10 times with a random 90/10 train/test split and reported over  $\pm$ a standard deviation.}
	\label{uc-table-msll}
	\centering
	\resizebox{\textwidth}{!}{
		\begin{tabular}{cccccccc}
			\toprule
			Dataset & $N$ & $D$ & RBF & ARD & ARD Mat\'ern & FKL-PB (separate) & FKL-PB (shared) \\
			\midrule
			challenger & 23 & 4 & 0.83 $\pm$ 1.085 & 0.91 $\pm$ 1.951 & 0.383 $\pm$ 0.778 & \textbf{-0.053 $\pm$ 0.192} & 0.216 $\pm$ 0.292 \\
			fertility & 100 & 9 & -0.049 $\pm$ 0.075 &\textbf{ -0.094 $\pm$ 0.137} & -0.077 $\pm$ 0.295 & 0.013 $\pm$ 0.06 & -0.0 $\pm$ 0.017 \\
			concreteslump & 103 & 7 & 30.821 $\pm$ 12.039 & 20.055 $\pm$ 11.079 & 17.247 $\pm$ 9.789 & -0.125 $\pm$ 0.131 & \textbf{-2.57 $\pm$ 0.23} \\
			servo & 167 & 4 & -1.076 $\pm$ 0.216 & -1.242 $\pm$ 0.386 & -1.25 $\pm$ 0.121 & \textbf{-1.28 $\pm$ 0.218} & -0.981 $\pm$ 0.272 \\
			yacht & 308 & 6 & 5.136 $\pm$ 8.696 & -2.001 $\pm$ 2.369 & 4.943 $\pm$ 7.521 & \textbf{-2.62 $\pm$ 0.225} & -2.477 $\pm$ 0.17 \\
			autompg & 392 & 7 & -1.065 $\pm$ 0.216 & -0.93 $\pm$ 0.306 & \textbf{-1.085 $\pm$ 0.152} & -1.034 $\pm$ 0.149 & -0.888 $\pm$ 0.482 \\
			boston & 506 & 13 & -0.912 $\pm$ 0.196 & -1.077 $\pm$ 0.213 & -1.031 $\pm$ 0.13 & -0.86 $\pm$ 0.085 &\textbf{ -1.191 $\pm$ 0.109} \\
			stock & 536 & 11 & -0.831 $\pm$ 0.082 & -0.82 $\pm$ 0.088 &\textbf{ -0.868 $\pm$ 0.105} & 0.014 $\pm$ 0.04 & -0.001 $\pm$ 0.017 \\
			pendulum & 630 & 9 & -1.12 $\pm$ 0.084 & -1.358 $\pm$ 0.147 & -1.586 $\pm$ 0.227 & -0.323 $\pm$ 0.181 & \textbf{-1.685 $\pm$ 0.263} \\
			energy & 768 & 8 & -1.684 $\pm$ 0.127 & -3.062 $\pm$ 0.093 & -3.11 $\pm$ 0.05 & \textbf{-3.49 $\pm$ 0.133} & -3.302 $\pm$ 0.081 \\
			concrete & 1030 & 8 & -0.417 $\pm$ 0.232 & -0.717 $\pm$ 0.171 &\textbf{ -0.745 $\pm$ 0.154} & -0.489 $\pm$ 1.37 & -0.311 $\pm$ 1.345 \\
			airfoil & 1503 & 5 & -0.994 $\pm$ 0.064 & -1.177 $\pm$ 0.078 & -1.31 $\pm$ 0.048 & -1.448 $\pm$ 0.336 & \textbf{-1.586 $\pm$ 0.198} \\
			\bottomrule
		\end{tabular}
	}
\end{table}

\begin{table}
	\caption{UCI Regression Negative Log-likelihoods, comparisons are with RBF, ARD, and ARD Mat\'ern kernels, $N$ points $D$ input dimensions. We compare to separate latent GPs for each input dimension. Each of the experiments were conducted 10 times with a random 90/10 train/test split and reported over  $\pm$ a standard deviation.}
	\label{uc-table-nll}
	\centering
	\resizebox{\textwidth}{!}{
		\begin{tabular}{cccccccc}
			\toprule
			Dataset & $N$ & $D$ & RBF & ARD & ARD Mat\'ern & FKL-PB (separate) & FKL-PB (shared) \\
			\midrule
			challenger & 23 & 4 & 5.74 $\pm$ 4.547 & 6.064 $\pm$ 7.283 & 3.753 $\pm$ 3.05 &\textbf{ 2.82 $\pm$ 0.809} & 2.966 $\pm$ 0.854 \\
			fertility & 100 & 9 & -3.901 $\pm$ 1.76 & -2.861 $\pm$ 2.187 & \textbf{-4.408 $\pm$ 2.582} & -1.83 $\pm$ 3.336 & -2.738 $\pm$ 1.252 \\
			concreteslump & 103 & 7 & 400.451 $\pm$ 134.157 & 282.544 $\pm$ 124.796 & 250.299 $\pm$ 108.762 & 60.248 $\pm$ 2.542 & \textbf{33.016 $\pm$ 1.965} \\
			servo & 167 & 4 & 5.144 $\pm$ 3.995 & 1.101 $\pm$ 5.871 & 1.374 $\pm$ 3.14 & \textbf{0.93 $\pm$ 3.867} & 4.686 $\pm$ 5.271 \\
			yacht & 308 & 6 & 221.42 $\pm$ 271.437 & 1.65 $\pm$ 76.479 & -\textbf{19.949 $\pm$ 14.092} & -15.703 $\pm$ 8.233 & -14.52 $\pm$ 4.7 \\
			autompg & 392 & 7 & 96.189 $\pm$ 8.025 & 104.563 $\pm$ 13.36 & \textbf{94.012 $\pm$ 5.033} & 98.942 $\pm$ 6.135 & 101.757 $\pm$ 19.333 \\
			housing & 506 & 13 & 139.617 $\pm$ 11.546 & 131.22 $\pm$ 15.034 & 130.841 $\pm$ 10.506 & 143.75 $\pm$ 5.714 & \textbf{122.618 $\pm$ 3.91} \\
			stock & 536 & 11 & \textbf{-191.624 $\pm$ 1.626 }& -191.515 $\pm$ 1.472 & -191.154 $\pm$ 1.318 & -140.055 $\pm$ 6.679 & -147.805 $\pm$ 2.577 \\
			pendulum & 630 & 9 & 84.964 $\pm$ 3.402 & 69.371 $\pm$ 7.299 & 62.64 $\pm$ 5.692 & 141.121 $\pm$ 20.914 & \textbf{53.86 $\pm$ 16.301} \\
			energy & 768 & 8 & 157.1 $\pm$ 8.894 & 52.118 $\pm$ 5.835 & 47.776 $\pm$ 3.591 & \textbf{17.808 $\pm$ 9.927} & 30.222 $\pm$ 6.881 \\
			concrete & 1030 & 8 & 395.596 $\pm$ 21.02 & 361.792 $\pm$ 20.077 & \textbf{357.248 $\pm$ 14.532} & 384.242 $\pm$ 140.779 & 405.779 $\pm$ 137.561 \\
			airfoil & 1503 & 5 & 358.932 $\pm$ 8.932 & 325.059 $\pm$ 6.605 & 305.588 $\pm$ 7.462 & 284.895 $\pm$ 48.796 & \textbf{270.073 $\pm$ 28.424} \\
			\bottomrule
		\end{tabular}

	}
\end{table}

\section{Large-Scale Precipitation Extrapolation}
\label{sec: big-prcp}

We demonstrate the scalability and practicality of FKL by extending this to a much larger dataset; modeling 108 different stations in seven American states across the northeast (ME, MA, VT, NH, RI, CT, NY) with a single latent Gaussian process, training on the first 300 days of the year, and attempting to extrapolate on the final 65 days. Despite not including any geographic information (e.g. longitude and latitude), FKL fits the trends across this climatologically diverse region. We show extrapolation on 120 stations in Figure 15 in the Appendix. Note that this corresponds to a dataset size of greater than 30,000 data points, and that we were able to fit this dataset on a single Nvidia 1080 Ti GPU in roughly 30 minutes. 

Each blue curve corresponds to the predictive \emph{mean} of a single component in the mixture of Gaussians predictive distribution. The shade corresponds to the 2 stdev uncertainty associated with each component. This representation shows the different Gaussian distributions that come together to form the unconditional predictive distributions.

\begin{figure}

	\resizebox{\textwidth}{!}{
		\includegraphics[width=2.7in]{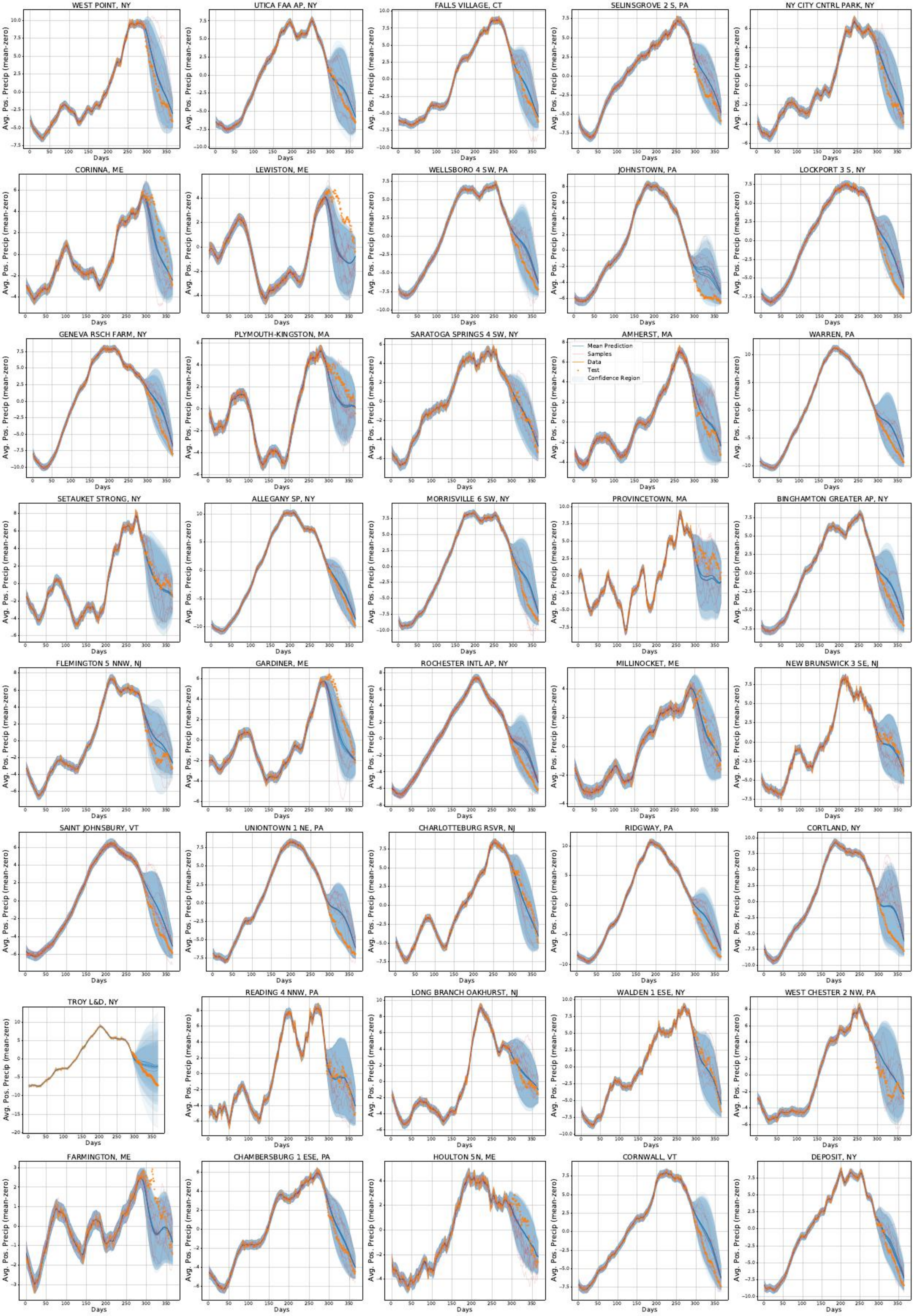}
	}
	\caption{40 stations modelled in the multi-task extrapolation test. The multi-task FKL both interpolates and extrapolates well even for relatively geographically diverse datasets. }
	\label{tab:precip_stations1}
\end{figure}

\begin{figure}

	\resizebox{\textwidth}{!}{
			\includegraphics[width=2.7in]{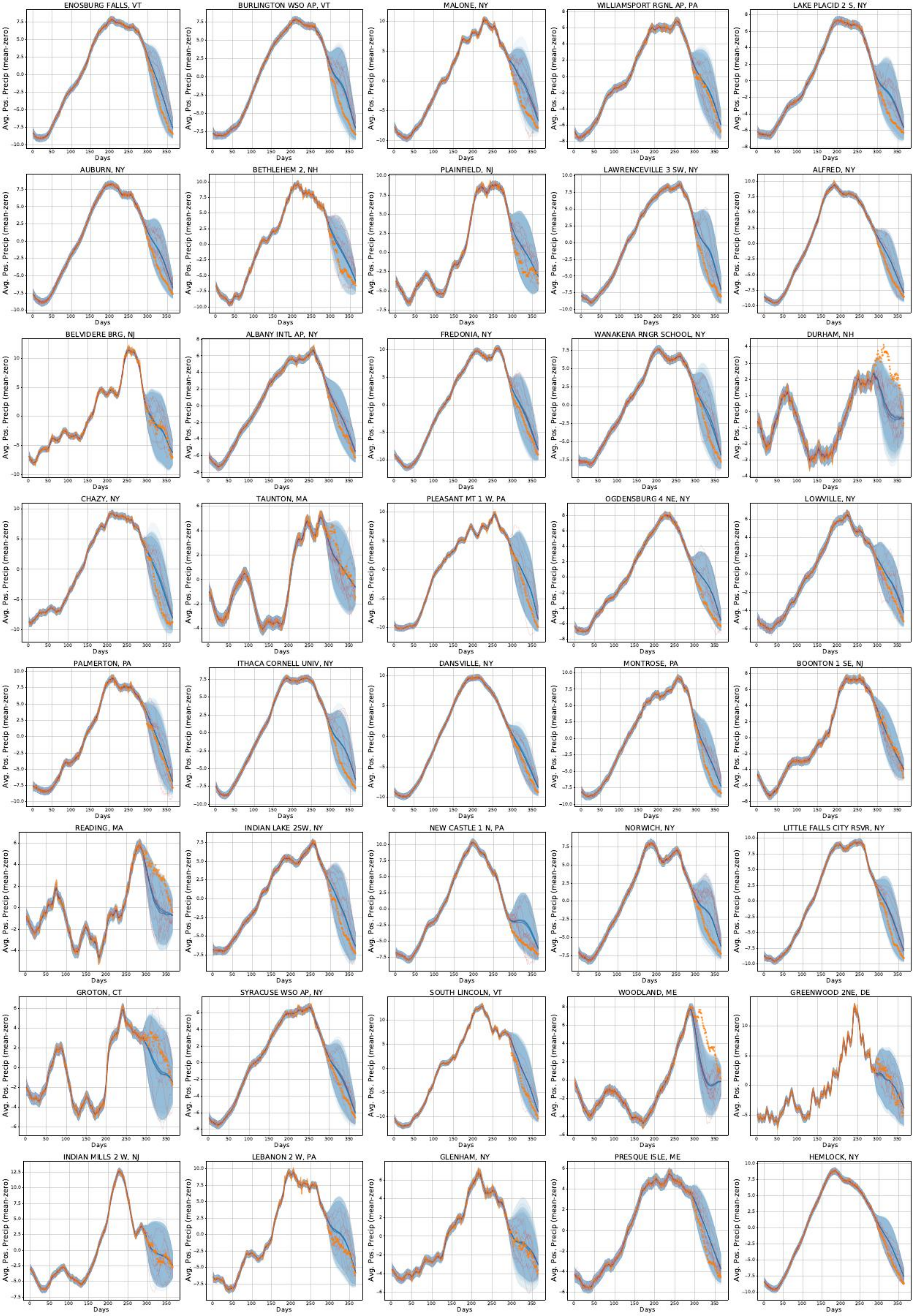}
	}
	\caption{40 stations modelled in the multi-task extrapolation test. The multi-task FKL both interpolates and extrapolates well even for relatively geographically diverse datasets. }
	\label{tab:precip_stations2}
\end{figure}

\begin{figure}
	\resizebox{\textwidth}{!}{
		\includegraphics[width=2.7in]{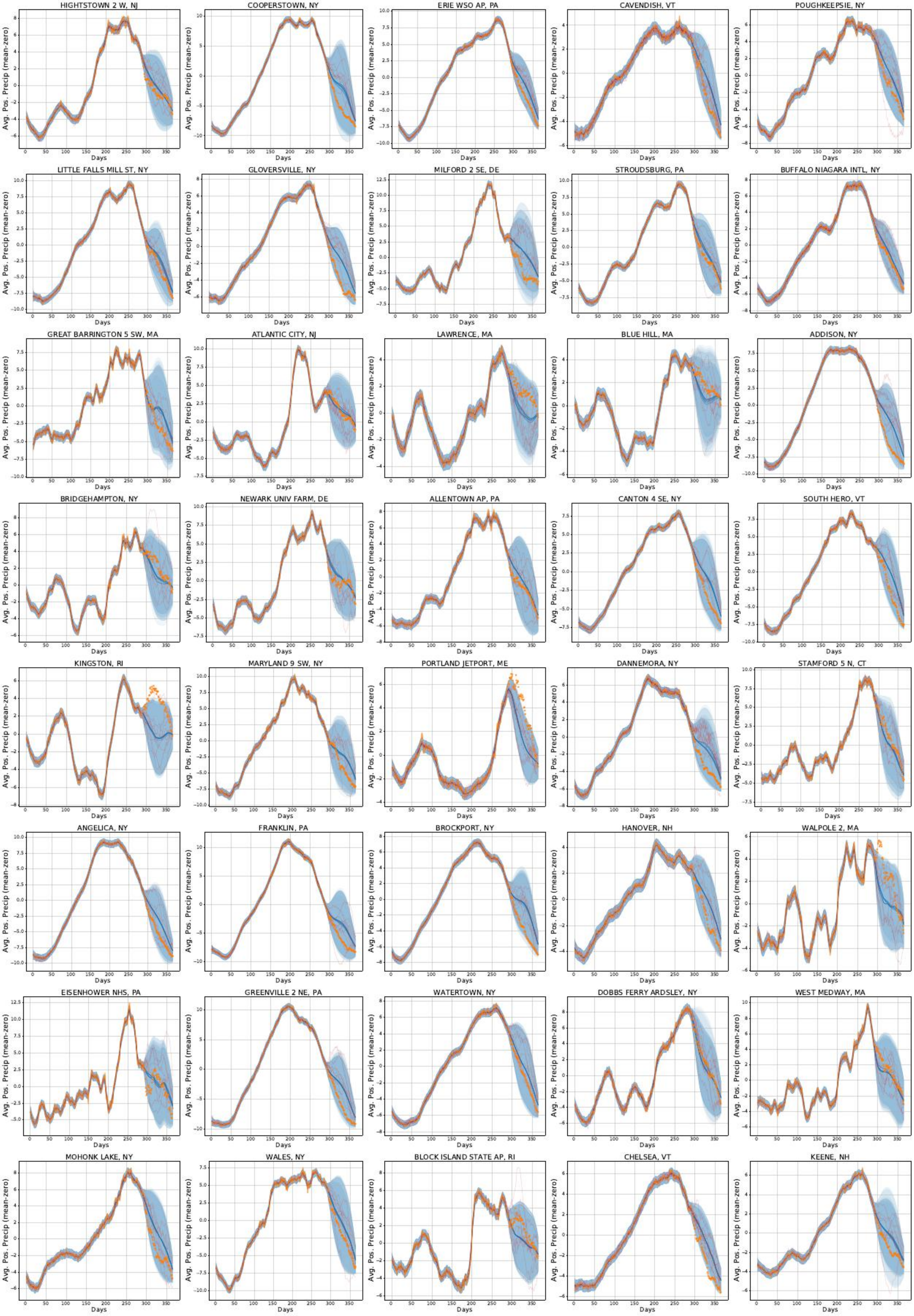}
	}
	\caption{40 stations modeled in the multi-task extrapolation test. The multi-task FKL both interpolates and extrapolates well even for relatively geographically diverse datasets. }
	\label{tab:precip_stations3}
\end{figure}

\begin{figure}
	\resizebox{\textwidth}{!}{
		\includegraphics[width=2.7in]{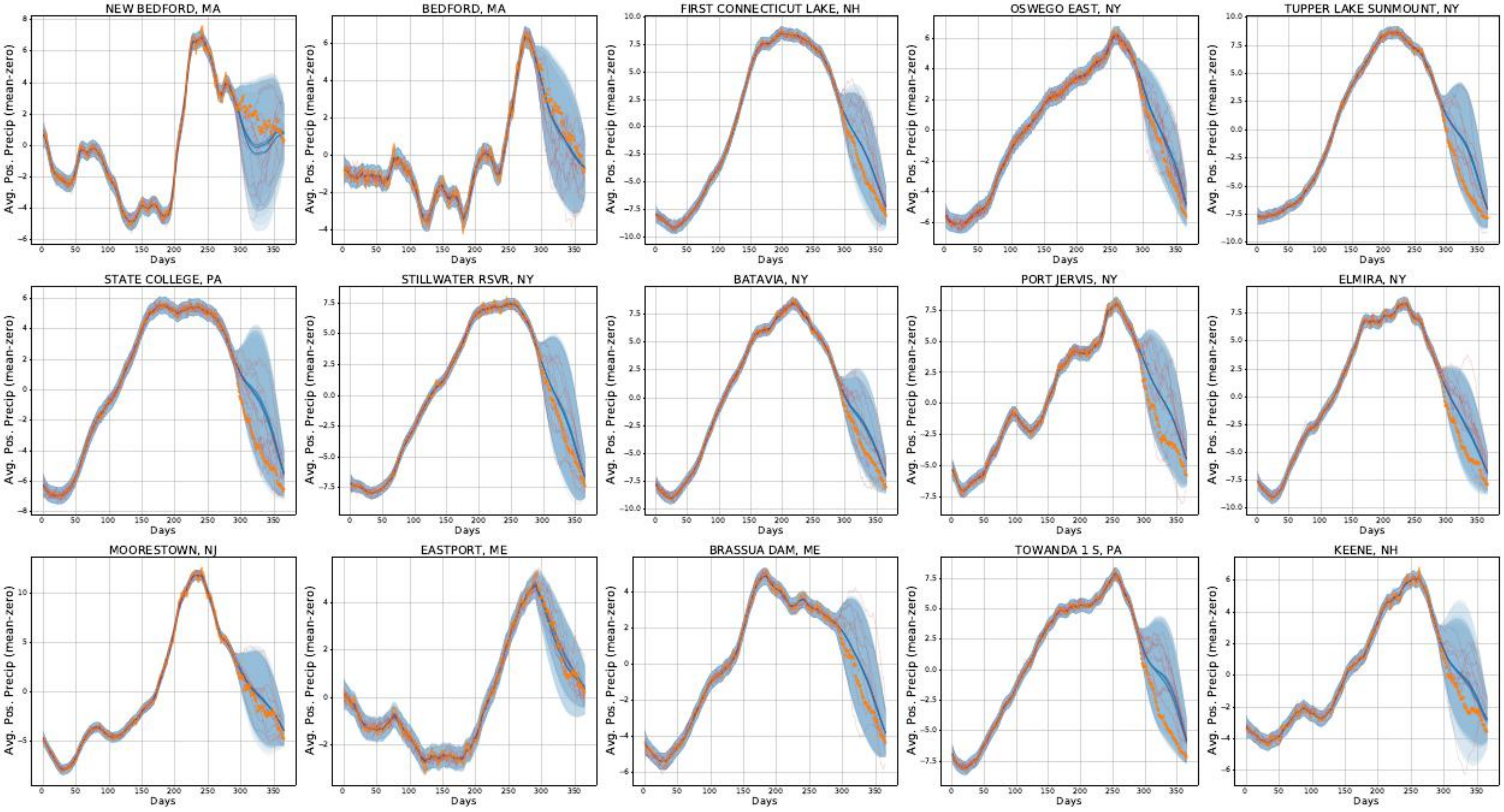}
	}
	\caption{15 stations modeled in the multi-task extrapolation test. The multi-task FKL both interpolates and extrapolates well even for relatively geographically diverse datasets. }
	\label{tab:precip_stations4}
\end{figure}

\begin{figure}
	\includegraphics[width=\textwidth]{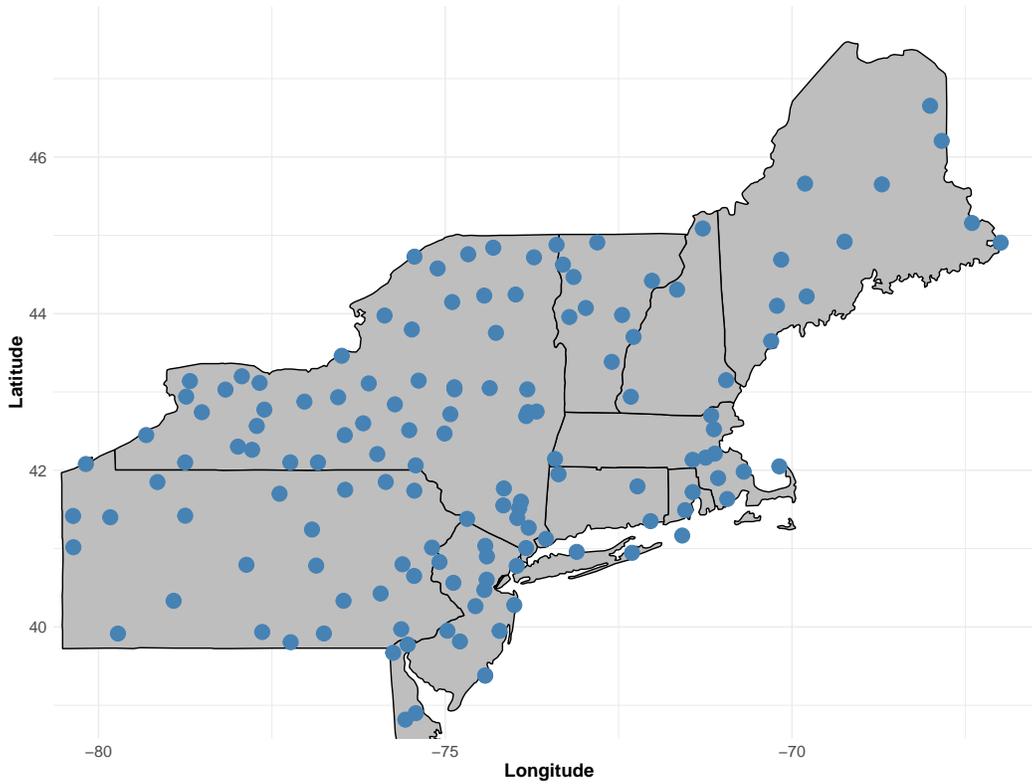}
	\caption{Map of locations used for large scale multi task experiment.}
\end{figure}

\end{document}